\documentclass[journal]{IEEEtran}

\usepackage{url}
\usepackage{graphicx}
\usepackage{subfigure}
\usepackage{caption}
\usepackage{cite}
\usepackage[ruled]{algorithm2e}
\usepackage{mathrsfs}
\usepackage{amsmath}
\usepackage{amssymb}
\usepackage{booktabs}
\usepackage{colortbl}
\usepackage{multirow}
\usepackage{amsthm}

\begin{document}

\title{Efficient Federated Learning for AIoT Applications Using Knowledge Distillation}

\author{Tian~Liu, Zhiwei~Ling, Jun~Xia, Xin~Fu,~\IEEEmembership{~Senior Member,~IEEE},
Shui~Yu,~\IEEEmembership{~Senior Member,~IEEE},
and Mingsong~Chen,~\IEEEmembership{~Senior Member,~IEEE}

\IEEEcompsocitemizethanks{

\IEEEcompsocthanksitem Tian Liu, Zhiwei Ling, Jun Xia and Mingsong Chen are with the 
MoE Engineering Research Center of Software/Hardware Co-design Technology and Application,
East China Normal University,
Shanghai, 200062, China (email: liutian2534@qq.com, \{zwling, jxia, mschen\}@sei.ecnu.edu.cn). 
Tian Liu is also with the Department of Information Science and Engineering,
Zaozhuang University, Zaozhuang 277160, China.
Xin Fu is with the Department of Electrical and Computer
Engineering, University of Houston, Houston, TX 77204 USA (e-mail: xfu8@central.uh.edu).
Shui Yu is with the School of Computer Science, University of Technology Sydney, Australia (e-mail: shui.yu@uts.edu.au).
}}


\maketitle

\begin{abstract}
As a promising distributed machine learning paradigm, Federated Learning (FL) trains a central model with decentralized data without compromising user privacy, which makes it widely used by Artificial Intelligence Internet of Things (AIoT) applications. However, the traditional FL suffers from model inaccuracy, since it trains local models only using hard labels of data while useful information of incorrect predictions with small probabilities is ignored. Although various solutions try to tackle the bottleneck of the traditional FL, most of them introduce significant communication overhead, making the deployment of large-scale AIoT devices a great challenge. To address the above problem, this paper presents a novel Distillation-based Federated Learning (DFL) method that enables efficient and accurate FL for AIoT applications. By using Knowledge Distillation (KD), in each round of FL  training, our approach uploads both the soft targets and local model gradients to the cloud server for aggregation, where the aggregation results are then dispatched to AIoT devices for the next round of local training. During the DFL local training, in addition to hard labels, the model predictions approximate soft targets, which can improve model accuracy by leveraging the knowledge of soft targets. To further improve our DFL model performance, we design a dynamic adjustment strategy of loss function weights for tuning the ratio of KD and FL, which can maximize the synergy between soft targets and hard labels. Comprehensive experimental results on well-known benchmarks show that our approach can significantly improve the model accuracy of FL without introducing significant communication overhead.
\end{abstract}

\begin{IEEEkeywords}
AIoT, dynamic adjustment strategy, federated learning, knowledge distillation, model accuracy.
\end{IEEEkeywords}

\IEEEpeerreviewmaketitle

\section{Introduction}\label{sec:intro}

\IEEEPARstart{A}{long} with the proliferation of Artificial Intelligence (AI) and Internet of Things (IoT), Federated Learning (FL)~\cite{communication,largescale,towards} techniques are increasingly used in safety-critical AI IoT (AIoT) applications (e.g., autonomous driving, commercial surveillance, and industrial control~\cite{deep,when}).
Different from centralized machine learning, FL enables keeping data samples distributed while sharing the sample knowledge among all the AIoT devices.
In FL, the cloud server is responsible for dispatching and aggregating model gradients rather than collecting samples from AIoT devices through the network, which can greatly reduce the communication overhead and protect the data privacy of AIoT devices during the model training process.

Although FL enables effective collaboration among AIoT devices and the cloud server, it drastically suffers from its model inaccuracy caused by the loss of knowledge during model training~\cite{dist}.
The optimization objective of FL local training is to minimize the distance between the correct prediction and the hard label and ignore all the incorrect predictions~\cite{communication}.
However, the ignoring of incorrect predictions results in the loss of knowledge since the knowledge is a learned mapping from input vectors to output vectors, and all the sample-to-prediction mappings are part of the knowledge according to~\cite{dist,distonline}.
The probability of incorrect predictions represents the similarities between the current sample and other different categories.
Therefore, the traditional FL based on hard labels loses some knowledge during the model training process, resulting in decreased FL model accuracy.

Since Knowledge Distillation (KD) can enhance the model knowledge and the model generalization ability, it is used to improve the model accuracy~\cite{dist}.
During the ``student model'' training process, there are two optimization objectives, i.e., hard labels of data and soft targets from the ``teacher model''.
The loss function of the ``student model'' is defined as the sum of the cross-entropy loss function (i.e., the distance between model predictions and the corresponding hard labels of data) and the Kullback-Leibler divergence loss function (i.e., the distance between model predictions and the corresponding soft targets from the ``teacher model''). 
As an online paradigm of KD, Federated Distillation (FD) implements collaborative training of different device models only by interacting soft targets between the cloud server and all the devices~\cite{cefd,fd,feded}.
However, all these methods focus more on the fundamental problems of network resource limitation for large-scale architecture rather than the FL performance improvement.

To improve the FL model accuracy, various methods have been proposed, e.g., global control variable-based methods~\cite{scaffold,crosssilo}, reinforcement learning-based methods~\cite{reinforcement}, device grouping-based methods~\cite{multicenter,hierarchical}, and KD-based methods~\cite{ensembledist}.
However, all these mentioned methods improve FL performance using complex reinforcement learning strategies or global variables with large sizes.
Therefore, most of them are unsuitable for AIoT applications with limited network and memory resources.
Moreover, these KD-based methods require collecting data distribution and sample categories from all devices or constructing public datasets, which brings huge risks to data privacy protection.
Therefore, how to design an efficient and accurate FL without introducing significant communication overhead and ensuring data privacy is becoming a great challenge in AIoT design.

In order to address the above challenges, this paper presents a novel Distillation-based FL method named DFL that can effectively enhance the model knowledge during the FL training process.  
Unlike the traditional FL that only trains models based on hard labels of device samples, our proposed DFL method set two optimization objectives for the model, i.e., the hard labels of data samples and the corresponding soft targets.
We aggregate label-wise sample logits as the soft targets of the ``teacher model'' and dispatch them together with the global model for FL model training, which introduces negligible extra network overhead as the soft target size is always much smaller than the global model.
In this way, our DFL method can increase the model accuracy by incorporating the knowledge of soft targets into the model training.
This paper makes the following three major contributions:
\begin{itemize}
	\item To improve the model accuracy of DFL, we present a novel architecture that combines the merits of both global soft targets and model gradients for the purpose of knowledge enhancement.
	\item To wisely utilize the knowledge represented by soft targets, we design a dynamic adjustment strategy, which can tune the ratio of loss functions of soft targets and hard labels during the DFL training.
	\item We conduct both theoretical and empirical analysis on the convergence of DFL and prove that DFL converges as fast as FedAvg in arbitrarily heterogeneous data scenarios.
\end{itemize}

We implement our approach using our proposed DFL architecture and the dynamic adjustment strategy. Comprehensive experimental results show that our proposed approach can achieve better performance than state-of-the-art methods without introducing drastic communication overhead.

The rest of this paper is organized as follows. 
After the introduction to related works in Section~\ref{sec:related}, Section~\ref{sec:framework} gives the details of our DFL approach. 
Section~\ref{sec:case} presents the experimental results, showing the effectiveness of our approach.
Finally, Section~\ref{sec:conclusions} concludes the paper.

\section{Related Work}\label{sec:related}

As more and more safety-critical AIoT applications adopt FL, the FL model accuracy is becoming a major concern in AIoT design.
To improve the model accuracy, Hinton et al.~\cite{dist} proposed Kederated Distillation (KD) to enhance the model knowledge with soft targets.  
To apply the benefits of KD to AIoT applications, various online versions of KD have been investigated.
For example, Anil et al.~\cite{distonline} proposed the co-distillation method with data samples shared by all the AIoT devices.
Based on FD and federated data augmentation, Jeong et al.~\cite{fd} used Generative Adversarial Networks (GAN)~\cite{gan} to generate a public dataset and carried out KD on the public dataset during the model training process.
By leveraging an unlabeled public dataset, Itahara et al.~\cite{dbss} proposed a distillation-based semi-supervised FL algorithm that exchanges outputs of local models among mobile devices.
However, all these FD approaches above focus on reducing communication overhead rather than improving model accuracy.
Moreover, these methods with public datasets introduce risks of privacy exposure that cannot be ignored.

In order to improve the FL model inference accuracy, various methods have been investigated.
For example, Karimireddy et al.~\cite{scaffold} proposed a method named SCAFFOLD, using global control variables to correct the ``client-drift'' in the local training process.
Similar to SCAFFOLD, Huang et al.~\cite{crosssilo} presented a method employing the federated attentive message passing to promote more cooperation among similar devices. 
However, all the two methods upload/dispatch additional large-size controllers (i.e., the global control variables and the attentive messages) along with the model gradient between the cloud server and devices.
By using built-in generators, Zhu et al.~\cite{datafree} proposed a data-free KD approach named FedGen to address the problem of heterogeneous FL.
Lin et al.~\cite{ensembledist} proposed an ensemble distillation method that trains the central model with unlabeled data and the corresponding outputs of device models.
Nonetheless, these two methods are not feasible to deploy in real scenarios, since they require each device to upload their data distribution or sample categories, which brings the risk of data exposure and huge communication overhead.
Moreover, it is impractical to construct built-in generators or public datasets that are helpful for model training.
Therefore, the above methods are unsuitable for AIoT applications with data privacy requirements and limited network communication capabilities.

Although KD techniques are promising in enhancing the FL performance, their combination faces the aforementioned technical challenges.
Moreover, existing distillation approaches did not consider that the knowledge of soft targets changes with the model training process.  
Generally, the knowledge of soft targets increases with the number of training rounds since the model accuracy is improved.
To the best of our knowledge, our work is the first attempt that fully explores the synergy between the model gradients and global soft targets to further enable knowledge sharing among AIoT devices. 
Due to the enhanced knowledge obtained by soft targets using our proposed architecture and dynamic adjustment strategy, the accuracy of DFL models can be significantly improved, while the communication overhead is negligible.

\section{Our DFL Approach}\label{sec:framework}

\begin{figure*}[h]
	\centering
	\includegraphics[width=0.8\textwidth]{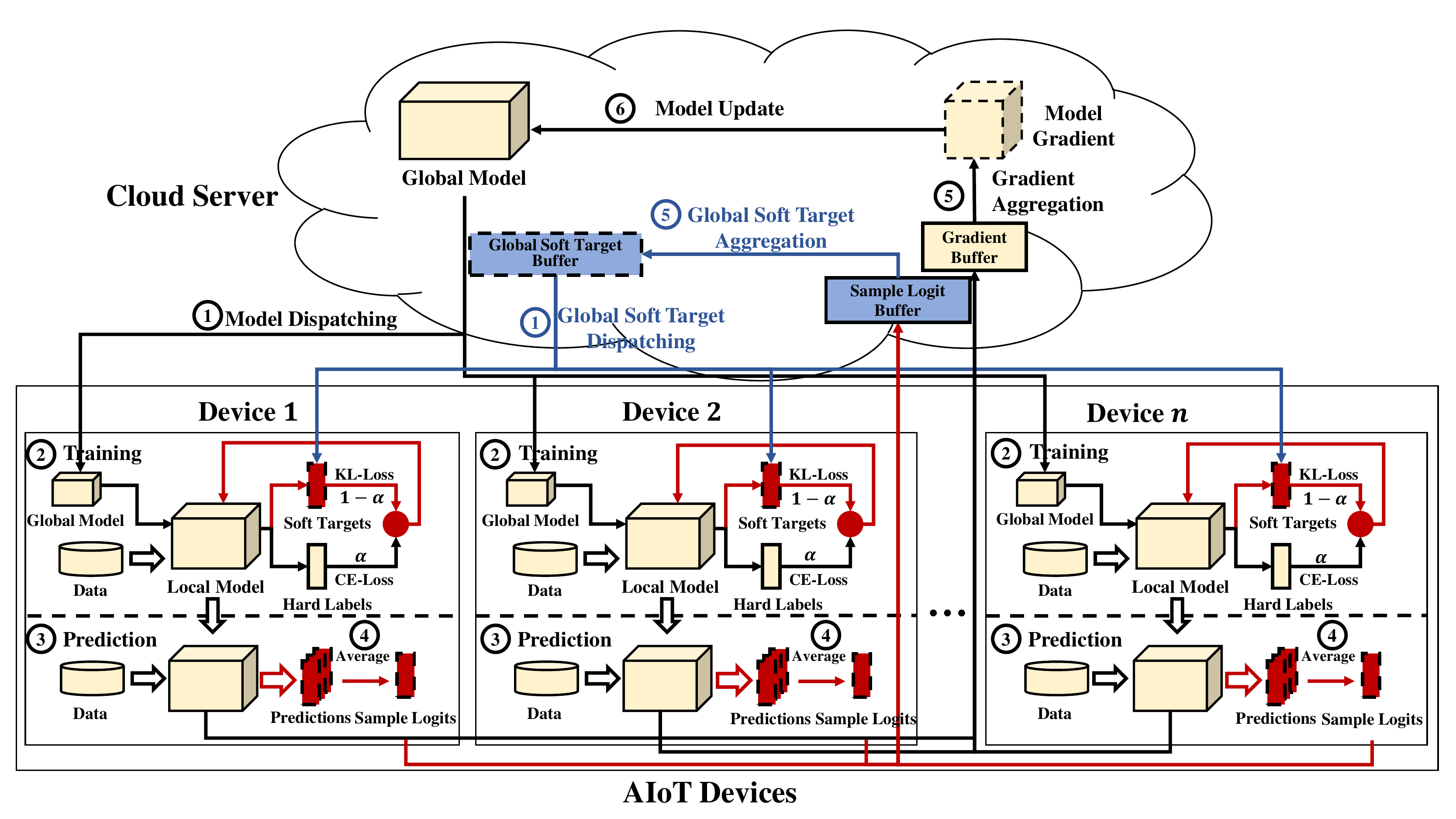}
	\caption{Architecture of our DFL approach}\label{architecture}
\end{figure*}

Typically, an AIoT application involves a cloud server and plenty of AIoT devices, where each AIoT device has limited communication and memory capacities.
In this paper, we focus on the model performance rather than the problem of incentive or fairness.
Therefore, we assume that at the beginning of our DFL architecture deployment, data samples are collected by each device and used for local model training.
The model to be trained is initially placed on the cloud server and dispatched to AIoT devices at the beginning of each training round.
Since soft targets can enhance the model knowledge and the model generalization ability~\cite{dist}, our approach introduces KD into our architecture to improve the model accuracy.
Unlike existing FD methods, our DFL approach uploads/dispatches model gradients and soft targets (generated using local samples in the previous round) simultaneously during the interaction between AIoT devices and the cloud server.
When a new device joins the AIoT application, it will receive the latest global model and soft targets from the cloud server and perfectly fit into the DFL model training.
The proposed DFL model training procedure is divided into two parts: i) the cloud server training part, which includes the dispatching, aggregation, and update of both model gradients and soft targets, and ii) the local training part, which trains local models using both local samples and dispatched soft targets.
Once the local training finishes, an AIoT device needs to figure out new label-wise sample logits for the following aggregation.
The following subsections will detail the key components and the convergence analysis of our DFL approach.

\subsection{Architecture of Our DFL Approach}

Figure~\ref{architecture} depicts the overall architecture of our proposed DFL approach, which is inspired by KD and FL methods.  
It mainly consists of a central cloud server and plenty of AIoT devices.  
To guarantee data privacy, data samples are collected by AIoT devices and cannot be shared with other devices and the cloud server.
As shown in Figure~\ref{architecture}, our DFL architecture has three parts, i.e., the FL processing part (marked in yellow), the Soft Target Processing Module (STPM) in the cloud server (marked in blue), and the STPM in AIoT devices (marked in red).
Our approach uses STPM in the cloud server to aggregate the label-wise sample logits as the global soft targets and dispatches them to selected AIoT devices.
The STPM in AIoT devices involves three functions: i) receiving the global soft targets from the cloud server; ii) performing local training using both the global soft targets and the hard labels of local samples; and iii) generating new label-wise sample logits using updated local models.
The details of the FL processing part are neglected here since they are similar to that of the traditional FL.
Note that designers can customize the models they need to train according to the requirements and available resources of AIoT applications.

\subsection{Training Procedure of DFL}

The model training procedure of our DFL approach consists of two parts, i.e., the cloud server procedure and the local update procedure.
At the very beginning of the model training, AIoT devices randomly collect a set of data samples and save them in their local memory for model training, while the cloud server initializes the global model and soft targets.
Similar to the classic FL method (i.e., FedAvg~\cite{communication}), our approach randomly selects a fraction of AIoT devices participating in each round of model training due to the limited network resources of real AIoT applications. 
The collaboration of the cloud server and AIoT devices of our proposed DFL method will be detailed in the following two subsections.

\subsubsection{Cloud Server Procedure}

When the model training of our DFL approach starts, the cloud server first dispatches the current global model and soft targets to the selected AIoT devices.
After receiving the latest model and soft targets, the selected AIoT devices will conduct several epochs of local training, respectively.
At the end of round $r$, we upload both model gradients and the newly generated label-wise sample logits of all the selected AIoT devices to the cloud server for aggregations using the following formulas:
\begin{equation}
    w_{r+1} = \ w_r +  \frac{\sum_{k=1}^{K} |D_k| \times \Delta_{r+1}^k}{{\sum_{k=1}^{K}} |D_k|}_.
\end{equation}
\begin{equation}
    \mathscr{Y}_{r+1} = \ \frac{\sum_{k=1}^{K} |D_k| \times \ \mathscr{Y}_{r+1}^k}{{\sum_{k=1}^{K}} |D_k|}_.
\end{equation}
where $w$ and $\mathscr{Y}$ represent the model weight and the label-wise soft targets, respectively. $K$ denotes the number of AIoT devices selected in each round of model training. $\Delta_{r+1}^k$ and $\mathscr{Y}_{r+1}^k$ indicate the model gradient and sample logits of device $k$ in round $r+1$, $|D_k|$ represents the size of the dataset contained by the $k^{th}$ AIoT device.

\begin{algorithm}[htbp] 
	\KwIn{
		i) $N$, \# of total AIoT devices\;
		\hspace{0.37in}		ii) $c$, fraction of devices on each round\;
		\hspace{0.33in}		iii) $R$, \# of training rounds\;
		\hspace{0.35in}		iv) $D=\{D_1,\cdots,D_N\}$, set of datasets\; 
 	}
		 {\bf 1.} $\text{Initialize}(w,\mathscr{Y})$\;
		 {\bf 2.} $K   \leftarrow \text{Max}(c \cdot N, 1)$\;
		\For{$r\leftarrow$  1 to $R$}{
		    
		    {\bf 3.} $S   \leftarrow $ random set of $K$ devices\;
			{\bf 4.} $\text{Dispatch}(w_r, \mathscr{Y}_r, S)$\;
			\For{$ each \ device \ k \in S$}{
				{\bf 5.} $(\Delta_{r+1}^k, \mathscr{Y}_{r+1}^k)   \leftarrow \text{DeviceUpdate}(w_{r}, \mathscr{Y}_r)$\;
			}
			{\bf 6.} {$  w_{r+1} \ =  \ w_r +  \frac{\sum_{k=1}^{K} |D_{Idx(S_k)}| \times \Delta_{r+1}^k}{{\sum_{k=1}^{K}} |D_{Idx(S_k)}|}$}\;
			{\bf 7.} {$  \mathscr{Y}_{r+1} \ = \ \frac{\sum_{k=1}^{K} |D_{Idx(S_k)}| \times \ \mathscr{Y}_{r+1}^k}{{\sum_{k=1}^{K}} |D_{Idx(S_k)}|}$}\;
		}
	\caption{Cloud Server Procedure of DFL}
	\label{alg:1}
\end{algorithm}

Algorithm \ref{alg:1} shows the key steps involved in our DFL algorithm.
Step 1 initializes the global model and soft targets with $w$ and $\mathscr{Y}$ using the function \textit{Initialize}.
In step 2, we calculate the number of selected AIoT devices $K$ participating in each round of model training with the function \textit{Max}, where $C$ and $N$ denote the fraction and the number of total AIoT devices.
At the beginning of round $r$, step 3 randomly selects the devices participating in the model training of round $r$, where $S$ is used to save the selected devices.
Step 4 dispatches the global model and soft targets to all the selected devices in $S$.
In step 5, all the selected devices upload both model gradients and the newly generated label-wise sample logits to the cloud server.
Once the cloud server receives model gradients and label-wise sample logits from all the selected devices, steps 6-7 perform the aggregation.

\subsubsection{Local Update Procedure}\label{localupdate}

When selected AIoT devices receive the latest global model $w$ and soft targets $\mathscr{Y}$, they conduct the local update procedure.
The local update procedure of our DFL approach involves two stages, i.e., the local training stage and the new label-wise sample logit generation stage.
Similar to FedAvg, the predictions of our DFL model approximate the hard labels of local samples.
To further improve the model accuracy, our approach makes the model predictions approximate to the soft targets related to the corresponding hard labels as well.
Therefore, to make wisely use of the knowledge of both hard labels and soft targets, we design our loss function in model training as follows:
\begin{equation}
    \mathcal{L}(w) = \rho \ \mathcal{F}(y|w, Y)+(1-\rho) \ \mathcal{G}(y|w, \mathscr{Y}).
    \label{equ:3}
\end{equation}
where $\mathcal{F}(y|w, Y)$ denotes the cross-entropy loss function, which is the distance between the prediction $y$ and the corresponding hard label $Y$ of the sample.
$\mathcal{G}(y|w, \mathscr{Y})$ indicates the Kullback-Leibler divergence loss function, which is the distance between the prediction $y$ and the corresponding label-wise sample logits $\mathscr{Y}$ (extract from the global soft targets) of the sample.
The hyperparameter $\rho$ ($\rho \in [0,1]$) is the ratio of the two loss functions (see Section~\ref{sec:ratio}).
Since the objective of local training is to minimize the loss function $\mathcal{L}(w)$, we can get the model update for each epoch as follows:
\begin{equation}
    w =  w -  \eta \nabla \mathcal{L}(w).
    \label{equ:4}
\end{equation}
where $\eta$ denotes the learning rate and $\nabla$ indicates the gradient.
When the local training stage finishes, the new label-wise sample logit generation stage will be implemented.
All the updated models perform predictions with local samples and calculate the label-wise sample logits.
To improve communication efficiency, all the selected AIoT devices upload their model gradients (i.e., $\Delta$) rather than the updated models to the cloud server for aggregation at the end of each round.

\begin{equation}
    \Delta_{r+1}^k = w_{r+1}^k - w_{r}^k.
\end{equation}

Algorithm~\ref{alg:2} presents the local update process of our DFL in detail.
In steps 1-2 of the algorithm, AIoT devices receive the global model $w$ and soft targets $\mathscr{Y}$ from the cloud server and save the received global model.
Steps 3-6 show the implementation of the local training stage.
At the beginning of each local epoch, step 3 makes predictions of data samples with the local model using the function \textit{Prediction}.
Step 4 calculates the current ratio of the two-loss functions $\rho$ with the index $r$ and the number $R$ of training rounds and the threshold $\mathcal{T}$.
Steps 5-6 iteratively update the model weight, where the loss function is defined in Formulas~\ref{equ:3}.
After the local training stage finishes, steps 7-9 generate the label-wise sample logits.
To save network resources, step 9 calculates the model gradient for upload.
Finally, step 10 uploads the model gradient and label-wise sample logits to the cloud server for aggregation.

\begin{algorithm}[h] 
	
	\KwIn{
		i) $E$, \# of local epochs\;
		\hspace{0.38in}		ii) $D$, device dataset with hard labels $Y$\; 
		\hspace{0.35in}		iii) $\eta$, learning rate\;
		\hspace{0.35in}		iv) $R$, \# of total communication rounds\;
        \hspace{0.38in}		v) $\mathcal{T}$, threshold of the loss function ratio\;
	}
	    \textbf{DeviceUpdate($w, \mathscr{Y}$):}\\
	    {\bf 1.} $\text{Receive} \ (w, \mathscr{Y}) \ \text{from the cloud server}$\;
		{\bf 2.} $temp  \ = \ w$\;
		\For{$e\leftarrow$  1 to $E$}{
		    
		    {\bf 3.} $y   \leftarrow \text{Prediction}(w, D)$\;
		    {\bf 4.} $\rho   \leftarrow \text{Max}(1-\frac{r}{R}, \mathcal{T})$\;
		    {\bf 5.} $\mathcal{L}(w) = \rho \ \mathcal{F}(y|w, Y)+(1-\rho) \ \mathcal{G}(y|w, \mathscr{Y})$\;
			{\bf 6.} {$ w =  w -  \eta \nabla \mathcal{L}(w)$}\;
		}
		{\bf 7.} $y \  \leftarrow \ \text{Prediction}(w, D)$\;
        {\bf 8.} $\mathscr{Y}   \leftarrow \ \text{LabelWiseAverage}(y)$\;
		{\bf 9.} $\Delta  \ = \  w-temp$\;
        {\bf 10.} $\text{Send} \ (\Delta, \mathscr{Y}) \ \text{to the cloud server}$\;
	\caption{Local Update Procedure of DFL} 
	\label{alg:2}
\end{algorithm}

\subsection{Dynamic Adjustment Strategy}\label{sec:ratio}

The ratio of the two-loss functions plays an important role in DFL local training since the weight of soft targets greatly impacts the model training.
Generally, the knowledge of soft targets depends on the model accuracy, and the model accuracy increases as the number of training rounds increases.
Therefore, there is insufficient knowledge of soft targets in the early stage of the model training process since the model is randomly initialized.
In this case, the soft targets will make the model optimize in the wrong direction, which will slow down the model training.
The knowledge of soft targets increases as the training continues, which can enhance the model with the knowledge that the hard labels do not have.
However, the model training cannot rely mainly on soft targets according to~\cite{dist}.
Therefore, we need to set a threshold $\mathcal{T}$ to fix the ratio of the two-loss functions in the late stage of the model training process so that the model can achieve the best performance.
To maximize the use of the soft targets and reduce their side effects, we design a dynamic adjustment strategy to control the loss function ratio as follows:
\begin{equation}
    \rho  \ = \  Max(1-\frac{r}{R}, \mathcal{T}).
    \label{equ:6}
\end{equation}
where $r$ and $R$ represent the index of the current round and the total number of overall training rounds, respectively. $\mathcal{T}$ denotes the threshold to fix the ratio of the two-loss functions.
As shown in Formula~\ref{equ:6}, in the early stage of the model training process, the cross-entropy loss function is given a high proportion and gradually decreases with the number of training rounds, while the Kullback-Leibler divergence loss function is the opposite.  
The discussion about the optimal threshold is in the experimental part.

\subsection{Convergence Analysis of Our DFL Approach}\label{sec:convergence}
Inspired by~\cite{convergence}, we analyze the convergence rate of our DFL approach with two device participation scenarios (i.e., full device participation and partial device participation).
We define the distributed optimization model of our DFL approach as follows:
\begin{equation}
\mathop{\min}\limits_{w} \Big\{\Phi(w) \triangleq \sum_{k = 1}^N p_k (\mathcal{F}_k(w) + \mathcal{G}_k (w)) \Big\}.
\end{equation}
where $N$ is the total number of all the AIoT devices, $p_k$ is the probability of selecting the $k^{th}$ device such that $p_k \geq 0$ and $\sum_{k = 1}^N p_k = 1$. 
$\mathcal{F}_k(w)$ and $\mathcal{G}_k (w)$ are two loss functions (i.e., the cross-entropy loss function and the Kullback-Leibler divergence loss function) which are defined as follows:
\begin{equation}
\begin{split}
&\mathcal{F}_k (w) \triangleq \frac{1}{n_k} \sum_{j = 1}^{n_k} \mathcal{F} (w; x_{k,j}),\\
&\mathcal{G}_k (w) \triangleq \frac{1}{n_k} \sum_{j = 1}^{n_k} \mathcal{G} (w; x_{k,j}^{'}).
\end{split}
\end{equation}
where $n_k$ is the number of local samples in the $k^{th}$ device, $x_{k,j}$ is the local training samples concluding pictures and hard labels, $x_{k,j}^{'}$ is the combination of the local samples and its corresponding soft targets.

Similar to~\cite{convergence}, to analyze the convergence rate of our DFL approach, we make the following five assumptions on the functions $\mathcal{F}_1, \cdots, \mathcal{F}_N$ and $\mathcal{G}_1, \cdots, \mathcal{G}_N$.

\textbf{Assumption 1.} $\mathcal{F}_1,  \cdots, \mathcal{F}_N$ and $\mathcal{G}_1, \cdots, \mathcal{G}_N$ are all $L-smooth$: for all $v$ and $w$, $\mathcal{F}_k(v) \leq \mathcal{F}_k(w) + (v-w)^T \nabla \mathcal{F}_k(w) + \frac{L}{2} || v - w ||^2$, $\mathcal{G}_k(v) \leq \mathcal{G}_k(w) + (v-w)^T \nabla \mathcal{G}_k(w) + \frac{L}{2} || v - w ||^2$.

\textbf{Assumption 2.} $\mathcal{F}_1, \cdots, \mathcal{F}_N$ and $\mathcal{G}_1, \cdots, \mathcal{G}_N$ are all $\mu-strongly \ convex$: for all $v$ and $w$, $\mathcal{F}_k(v) \geq \mathcal{F}_k(w) + (v-w)^T \nabla \mathcal{F}_k(w) + \frac{\mu}{2} || v - w ||^2$, $\mathcal{G}_k(v) \geq \mathcal{G}_k(w) + (v-w)^T \nabla \mathcal{G}_k(w) + \frac{\mu}{2} || v - w ||^2$.

\textbf{Assumption 3.} Let $\xi_t^k$ and $\delta_t^k$ be sampled from the $k^{th}$ device’s local data uniformly at random. The variance of stochastic gradients in each device is bounded: $\mathbb{E}||\nabla \mathcal{F}_k (w_t^k, \xi_t^k) - \nabla \mathcal{F}_k (w_t^k) ||^2 \leq \alpha_k^2$ for $k = 1, \cdots, N$ and $\mathbb{E}||\nabla \mathcal{G}_k (w_t^k, \delta_t^k) - \nabla \mathcal{G}_k (w_t^k) ||^2 \leq \beta_k^2$ for $k = 1, \cdots, N$.

\textbf{Assumption 4.} The expected squared norm of stochastic gradients is uniformly bounded, i.e., $\mathbb{E}||\nabla \mathcal{F}_k (w_t^k, \xi_t^k) ||^2 \leq G_1^2$ and $\mathbb{E}||\nabla \mathcal{G}_k (w_t^k, \delta_t^k) ||^2 \leq G_2^2$ for all $k = 1, \cdots, N$ and $t = 1, \cdots, T - 1$.

\textbf{Assumption 5.} From the $t^{th}$ local SGD, the distribution of soft targets no longer changes, so that $\mathcal{G}_k (w)$ is the only dependent variable of $w$.

Based on the assumptions above, we first analyze the convergence rate of our DFL approach with full device participation.
The update of our DFL model can be described with the following formulas:
\begin{equation}\label{equ:9}
\begin{split}
&v_{t+1}^k = w_t^k - \eta_t (\nabla \mathcal{F}_k (w_t^k, \xi_t^k) + \nabla \mathcal{G}_k (w_t^k, \delta_t^k)), \\
&w_{t+1}^k = 
  \begin{cases}
    v_{t+1}^k, \qquad \qquad \qquad \qquad  if \quad T \nmid t + 1,\\
	\sum_{k=1}^N p_k v_{t+1}^k, \qquad \qquad \ if \quad T \mid t + 1.
  \end{cases}
\end{split}
\end{equation}
where $w_t^k$ is the local model parameter maintained in the $k^{th}$ device at the $t^{th}$ SGD step, $v_{t+1}^k$ is the immediate result of $w_t^k$ with one step of SGD update.
$T$ is the local SGD steps within one training round.
If $T \mid t + 1$, our DFL activates all the AIoT devices.
In our analysis, we define two virtual sequences:
\begin{equation}\label{equ:10}
\begin{split}
    \overline{v}_t = \sum_{k=1}^{N} p_k v_t^k, \qquad \overline{w}_t = \sum_{k=1}^{N} p_k w_t^k.
\end{split}
\end{equation}

By combining Formulas~\ref{equ:9} and~\ref{equ:10}, we always have $\overline{v}_t = \overline{w}_t$.
For convenience, we define $\overline{g}_t = \sum_{k=1}^{N} p_k \Big[ \nabla \mathcal{F}_k (w_t^k) + \nabla \mathcal{G}_k (w_t^k) \Big]$ and $g_t = \sum_{k=1}^{N} p_k \Big[ \nabla \mathcal{F}_k (w_t^k, \xi_t^k) + \nabla \mathcal{G}_k (w_t^k, \delta_t^k) \Big]$. 
Therefore, $\overline{v}_{t+1} = \overline{w}_t - \eta_t g_t$ and $\mathbb{E} [g_t] = \overline{g}_t$.
We have:
\begin{equation}
\begin{split}
&|| \overline{v}_{t+1} - w^\star ||^2 = || \overline{w}_t - \eta_t g_t - w^\star - \eta_t \overline{g}_t +\eta_t \overline{g}_t ||^2 \\
&= \underbrace{|| \overline{w}_t - w^\star - \eta_t \overline{g}_t ||^2}_{A_1} + \underbrace{2 \eta_t <\overline{w}_t - w^\star \eta_t \overline{g}_t, \overline{g}_t - g_t>}_{A_2} \\
&+ \underbrace{\eta_t^2 || g_t - \overline{g}_t ||^2}_{A_3}.
\end{split}
\end{equation}

Note that $\mathbb{E}A_2 = 0$. We next focus on bounding $A_1$. Again we divide $A_1$ into three terms:
\begin{equation}\label{equ:a11}
\begin{split}
A_1 &= || \overline{w}_t - w^\star - \eta_t \overline{g}_t ||^2 \\
&= || \overline{w}_t - w^\star ||^2 - \underbrace{2 \eta_t <\overline{w}_t - w^\star , \overline{g}_t>}_{B_1} + \underbrace{\eta_t^2 || \overline{g}_t ||^2}_{B_2}.
\end{split}
\end{equation}

We aim to bound $B_1$:
\begin{equation}\label{equ:b11}
\begin{split}
B_1 &= - 2 \eta_t <\overline{w}_t - w^\star , \overline{g}_t> \\
&= - 2 \eta_t \sum_{k = 1}^N p_k <\overline{w}_t - w^\star , \nabla \mathcal{F}_k (w_t^k) + \nabla \mathcal{G}_k (w_t^k)>. 
\end{split}
\end{equation}
where
\begin{equation}\label{equ:14}
\begin{split}
&<\overline{w}_t - w^\star , \nabla \mathcal{F}_k (w_t^k) + \nabla \mathcal{G}_k (w_t^k)> \\
&= <\overline{w}_t - w_t^k, \nabla \mathcal{F}_k (w_t^k)> + <w_t^k - w^\star , \nabla \mathcal{F}_k (w_t^k)> \\
&+ <\overline{w}_t - w_t^k, \nabla \mathcal{G}_k (w_t^k)> + <w_t^k - w^\star , \nabla \mathcal{G}_k (w_t^k)>.
\end{split}
\end{equation}

By Cauchy-Schwarz inequality and the inequality of Arithmetic and Geometric Means (AM-GM), we can get:
\begin{equation}\label{equ:15}
\begin{split}
&- <\overline{w}_t - w_t^k, \nabla \mathcal{F}_k(w_t^k)> \\
&\leq \frac{1}{2\eta_t} || \overline{w}_t - w_t^k ||^2 + \frac{1}{2}\eta_t || \nabla \mathcal{F}_k (w_t^k) ||^2 \\
&- <w_t^k - w^\star, \nabla \mathcal{F}_k(w_t^k)> \\
&\leq -(\mathcal{F}_k(w_t^k) - \mathcal{F}_k(w^\star)) - \frac{\mu}{2} || w_t^k - w^\star ||^2\\
&- <\overline{w}_t - w_t^k, \nabla \mathcal{G}_k(w_t^k)> \\
&\leq \frac{1}{2\eta_t} || \overline{w}_t - w_t^k ||^2 + \frac{1}{2}\eta_t || \nabla \mathcal{G}_k (w_t^k) ||^2 \\
&- <w_t^k - w^\star, \nabla \mathcal{G}_k(w_t^k)> \\
&\leq -(\mathcal{G}_k(w_t^k) - \mathcal{G}_k(w^\star)) - \frac{\mu}{2} || w_t^k - w^\star ||^2.
\end{split}
\end{equation}

Therefore, $B_1$ can be presented as Formula~\ref{equ:b12} based on Formulas~\ref{equ:b11},~\ref{equ:14}, and~\ref{equ:15}, i.e.,
\begin{equation}\label{equ:b12}
\begin{split}
B_1 &\leq \eta_t \sum_{k = 1}^N p_k (\frac{1}{\eta_t} || \overline{w}_t - w_t^k ||^2 + \eta_t || \nabla \mathcal{F}_k (w_t^k) ||^2 ) \\
&- 2 \eta_t \sum_{k = 1}^N p_k (\mathcal{F}_k (w_t^k) - \mathcal{F}_k (w^\star) + \frac{\mu}{2} || w_t^k - w^\star ||^2 ) \\
&+  \eta_t \sum_{k = 1}^N p_k (\frac{1}{\eta_t} || \overline{w}_t - w_t^k ||^2 + \eta_t || \nabla \mathcal{G}_k (w_t^k) ||^2 ) \\
&- 2 \eta_t \sum_{k = 1}^N p_k (\mathcal{G}_k (w_t^k) - \mathcal{G}_k (w^\star) + \frac{\mu}{2} || w_t^k - w^\star ||^2 ).
\end{split}
\end{equation}

By using Assumption 1, $\mathcal{F}_k (\cdot)$ and $\mathcal{G}_k (\cdot)$ can be bounded with the following formulas:
\begin{equation}\label{equ:smooth}
\begin{split}
&|| \nabla \mathcal{F}_k (w_t^k) ||^2 \leq 2 L (\mathcal{F}_k (w_t^k) - \mathcal{F}_k^\star),\\
&|| \nabla \mathcal{G}_k (w_t^k) ||^2 \leq 2 L (\mathcal{G}_k (w_t^k) - \mathcal{G}_k^\star).
\end{split}
\end{equation}

Consequently, $B_2$ can be bounded using Formula~\ref{equ:b2} based on the convexity of $||\cdot||^2$ and Formula~\ref{equ:smooth}:
\begin{equation}\label{equ:b2}
\begin{split}
B_2 &= \eta_t^2 || \overline{g}_t ||^2 \leq \eta_t^2 \sum_{k = 1}^N p_k || \nabla \mathcal{F}_k (w_t^k) + \nabla \mathcal{G}_k(w_t^k) ||^2 \\
&\leq 2\eta_t^2 \sum_{k = 1}^N p_k \Big[ || \nabla \mathcal{F}_k (w_t^k) ||^2 + ||\nabla \mathcal{G}_k(w_t^k)||^2 \Big] \\
&\leq 4 L \eta_t^2 \sum_{k = 1}^N p_k \Big[ ( \mathcal{F}_k (w_t^k) - \mathcal{F}_k^{\star}) + ( \mathcal{G}_k (w_t^k) - \mathcal{G}_k^{\star}) \Big].
\end{split}
\end{equation}

Therefore, $A_1$ can be presented as Formula~\ref{equ:a12} by combining Formulas~\ref{equ:a11},~\ref{equ:b12} and~\ref{equ:b2}, i.e.,
\begin{equation}\label{equ:a12}
\footnotesize
\begin{split}
&A_1 \leq (1-2 \mu \eta_t) || \overline{w}_t - w^\star ||^2 + 2 \sum_{k = 1}^N p_k || \overline{w}_t - w_t^k ||^2 \\
&\underbrace{+ 6 L \eta_t^2 \sum_{k = 1}^N p_k (\mathcal{F}_k (w_t^k) - \mathcal{F}_k^\star)  + 6 L \eta_t^2 \sum_{k = 1}^N p_k (\mathcal{G}_k (w_t^k) - \mathcal{G}_k^\star)}_C \\
&\underbrace{- 2 \eta_t \sum_{k = 1}^N p_k (\mathcal{F}_k (w_t^k) - \mathcal{F}_k(w^\star))  - 2 \eta_t \sum_{k = 1}^N p_k (\mathcal{G}_k (w_t^k) - \mathcal{G}_k(w^\star))}_C.
\end{split}
\end{equation}

We next aim to bound C. We define $\gamma_t = 2 \eta_t (1 - 3 L \eta_t)$, $\Gamma = \Phi^\star - \sum_{k = 1}^N p_k \mathcal{F}_k^\star - \sum_{k = 1}^N p_k \mathcal{G}_k^\star$.
We split $C$ into three terms:

\begin{equation}\label{equ:20}
\small
\begin{split}
C &= -2 \eta_t (1 - 3 L \eta_t) \sum_{k = 1}^N p_k (\mathcal{F}_k (w_t^k) - \mathcal{F}_k^\star) \\
&-2 \eta_t (1 - 3 L \eta_t) \sum_{k = 1}^N p_k (\mathcal{G}_k (w_t^k) - \mathcal{G}_k^\star) \\
&+ 2 \eta_t \sum_{k = 1}^N p_k (\mathcal{F}_k(w^\star) - \mathcal{F}_k^\star) + 2 \eta_t \sum_{k = 1}^N p_k (\mathcal{G}_k(w^\star) - \mathcal{G}_k^\star) \\
&= -\gamma_t \sum_{k = 1}^N p_k (\mathcal{F}_k (w_t^k) - \Phi^\star) - \gamma_t \sum_{k = 1}^N p_k (\mathcal{G}_k (w_t^k) - \Phi^\star) \\
&+ (2 \eta_t - \gamma_t) \sum_{k = 1}^N p_k (\Phi^\star - \mathcal{F}_k^\star - \mathcal{G}_k^\star) - \gamma_t \Phi^\star \\
&= \underbrace{-\gamma_t \sum_{k = 1}^N p_k (\mathcal{F}_k (w_t^k) - \Phi^\star) - \gamma_t \sum_{k = 1}^N p_k (\mathcal{G}_k (w_t^k) - \Phi^\star)}_D \\
&+ 6 L \eta_t^2 \Gamma - \gamma_t \sum_{k = 1}^N p_k \Phi^\star.
\end{split}
\end{equation}

To bound D, we have:
\begin{equation}\label{equ:21}
\small
\begin{split}
&\sum_{k = 1}^N p_k (\mathcal{F}_k (w_t^k) - \Phi^\star) \\
&= \sum_{k = 1}^N p_k (\mathcal{F}_k (w_t^k) - \mathcal{F}_k (\overline{w}_t)) + \sum_{k = 1}^N p_k (\mathcal{F}_k (\overline{w}_t) - \Phi^\star) \\
&\geq \sum_{k = 1}^N p_k <\overline{v} \mathcal{F}_k (\overline{w}_t), w_t^k - \overline{w}_t> + \sum_{k = 1}^N p_k \mathcal{F}_k(\overline{w}_t) - \Phi^\star \\
&\geq - \frac{1}{2} \sum_{k = 1}^N p_k \Big[\eta_t || \overline{v} \mathcal{F}_k (\overline{w}_t) ||^2 + \frac{1}{\eta_t} || w_t^k - \overline{w}_t ||^2 \Big] \\
&+ \sum_{k = 1}^N p_k \mathcal{F}_k(\overline{w}_t) -\Phi^\star \\
&\geq - \sum_{k = 1}^N p_k \Big[\eta_t L (\mathcal{F}_k (\overline{w}_t) - \mathcal{F}_k^\star) + \frac{1}{2\eta_t} || w_t^k - \overline{w}_t ||^2 \Big] \\
&+ \sum_{k = 1}^N p_k \mathcal{F}_k(\overline{w}_t) -\Phi^\star.
\end{split}
\end{equation}
where the first inequality of Formula~\ref{equ:21} is from the convexity of $\mathcal{F}_k$, the second inequality of Formula~\ref{equ:21} is from AM-GM inequality and the third inequality of Formula~\ref{equ:21} is from Formula~\ref{equ:smooth}.
We use the same method to bound terms related to $\mathcal{G}_k$. Then we have: 
\begin{equation}\label{equ:22}
\begin{split}
D &\leq \gamma_t \sum_{k = 1}^N p_k \Big[\eta_t L (\mathcal{F}_k (\overline{w}_t) - \mathcal{F}_k^\star) + \frac{1}{2\eta_t} || w_t^k - \overline{w}_t ||^2 \Big]\\
&- \gamma_t (\sum_{k = 1}^N p_k \mathcal{F}_k(\overline{w}_t) -\Phi^\star) \\
&+ \gamma_t \sum_{k = 1}^N p_k \Big[\eta_t L (\mathcal{G}_k (\overline{w}_t) - \mathcal{G}_k^\star) + \frac{1}{2\eta_t} || w_t^k - \overline{w}_t ||^2 \Big] \\
&- \gamma_t (\sum_{k = 1}^N p_k \mathcal{G}_k(\overline{w}_t) -\Phi^\star).
\end{split}
\end{equation}

Therefore, by combining Formulas~\ref{equ:20} and~\ref{equ:22}, we can get:
\begin{equation}
\begin{split}
C &\leq \gamma_t (\eta_t L - 1) \sum_{k = 1}^N p_k (\mathcal{F}_k (\overline{w}_t) - \Phi^\star) \\
&+ \frac{\gamma_t}{2 \eta_t} \sum_{k = 1}^N p_k ||w_t^k - \overline{w}_t ||^2 + \gamma_t \eta_t L \sum_{k = 1}^N p_k (\Phi^\star - \mathcal{F}_k^\star) \\
&+ \gamma_t (\eta_t L - 1) \sum_{k = 1}^N p_k (\mathcal{G}_k (\overline{w}_t) - \Phi^\star) \\
&+ \frac{\gamma_t}{2 \eta_t} \sum_{k = 1}^N p_k ||w_t^k - \overline{w}_t ||^2 + \gamma_t \eta_t L \sum_{k = 1}^N p_k (\Phi^\star - \mathcal{G}_k^\star) \\
&+ 6 L \eta_t^2 \Gamma - \gamma_t \sum_{k = 1}^N p_k \Phi^\star \\
&\leq \gamma_t (\eta_t L - 1) \sum_{k = 1}^N p_k (\mathcal{F}_k (\overline{w}_t) - \frac{1}{2}\Phi^\star) \\
&+ \gamma_t (\eta_t L - 1) \sum_{k = 1}^N p_k (\mathcal{G}_k (\overline{w}_t) - \frac{1}{2}\Phi^\star) + \frac{\gamma_t}{\eta_t} \sum_{k = 1}^N p_k ||w_t^k - \overline{w}_t ||^2 \\
&+ \gamma_t \eta_t L \sum_{k = 1}^N p_k (\frac{1}{2} \Phi^\star - \mathcal{F}_k^\star) + \gamma_t \eta_t L \sum_{k = 1}^N p_k (\frac{1}{2}\Phi^\star - \mathcal{G}_k^\star) \\
&+ 6 L \eta_t^2 \Gamma - \gamma_t (\eta_t L - 1) \Phi^\star + \gamma_t \eta_t L \Phi^\star - \gamma_t \Phi^\star \\
&= \gamma_t (\eta_t L - 1) \sum_{k = 1}^N p_k (\mathcal{F}_k (\overline{w}_t) + \mathcal{G}_k (\overline{w}_t) - \Phi^\star) \\
&+ \frac{\gamma_t}{\eta_t} \sum_{k = 1}^N p_k ||w_t^k - \overline{w}_t ||^2 + \gamma_t \eta_t L ( \Phi^\star - \mathcal{F}_k^\star - \mathcal{G}_k^\star) + 6 L \eta_t^2 \Gamma \\
&\leq 2 \sum_{k = 1}^N p_k || w_t^k - \overline{w}_t ||^2 + (6 L \eta_t^2 + \gamma_t \eta_t L) \Gamma\\
 &\leq 2 \sum_{k = 1}^N p_k || w_t^k - \overline{w}_t ||^2 + 8 L \eta_t^2 \Gamma.
\end{split}
\end{equation}
where in the last inequality, we use the following three facts: i) $\eta_t L - 1 \leq - \frac{3}{4} \leq 0$ and $\sum_{k = 1}^N p_k (\mathcal{F}_k (\overline{w}_t) + \mathcal{G}_k (\overline{w}_t) - \Phi^\star) = \Phi (\overline{w}_t) - \Phi^\star \geq 0$, ii) $\Gamma \geq 0$ and $6 L \eta_t^2 + \gamma_t \eta_t L \leq 8 \eta_t^2 L$, and iii) $\frac{\gamma_t}{2 \eta_t} \leq 1$.
Recalling the expression of $A_1$ and plugging $C$ into it, we have:
\begin{equation}\label{equ:a13}
\small
\begin{split}
A_1 \leq (1 - 2 \mu \eta_t) || \overline{w}_t - w^\star ||^2 + 4 \sum_{k = 1}^N p_k || \overline{w}_t - w_t^k ||^2 + 8 L \eta_t^2 \Gamma.
\end{split}
\end{equation}

The variance of the stochastic gradients $\mathcal{F}$ and $\mathcal{G}$ in device $k$ is bounded by $\alpha^2_k$ and $\beta^2_k$.
Consequently, we have:
\begin{equation}\label{equ:27}
\footnotesize
\begin{split}
&\mathbb{E} || g_t - \overline{g}_t ||^2 \\
&= \mathbb{E} \Bigg|\Bigg| \sum_{k=1}^{N} p_k (\nabla \mathcal{F}_k (w_t^k, \xi_t^k) + \nabla \mathcal{G}_k (w_t^k, \xi_t^k) - \nabla \mathcal{F}_k (w_t^k) - \nabla \mathcal{G}_k (w_t^k)) \Bigg|\Bigg|^2 \\
&= \mathbb{E} \Bigg|\Bigg| \sum_{k=1}^{N} p_k (\nabla \mathcal{F}_k (w_t^k, \xi_t^k) - \nabla \mathcal{F}_k (w_t^k) ) \\
&+ \sum_{k=1}^{N} p_k ( \nabla \mathcal{G}_k (w_t^k, \xi_t^k) - \nabla \mathcal{G}_k (w_t^k)) \Bigg|\Bigg|^2 \\
&= \sum_{k=1}^{N} p_k^2 \mathbb{E} || \nabla \mathcal{F}_k (w_t^k, \xi_t^k) - \nabla \mathcal{F}_k (w_t^k)) ||^2 \\
&+ \sum_{k=1}^{N} p_k^2 \mathbb{E} || \nabla \mathcal{G}_k (w_t^k, \delta_t^k) - \nabla \mathcal{G}_k (w_t^k)) ||^2 \\
&\leq \sum_{k=1}^{N} p_k^2 (\alpha_k^2 + \beta_k^2).
\end{split}
\end{equation}

Since our DFL requires a communication round each $T$ SGD steps. Therefore, for any $t \geq 0$, there exists a $t_0 \leq t$, such that $t - t_0 \leq T - 1$ and $w_{t_0}^k = \overline{w}_{t_0}$ for all $k = 1, 2, \cdots , N$. 
Based on the fact that $\eta_t$ is non-increasing and $\eta_{t_0} \leq 2 \eta_t$ for all $t - t_0 \leq T - 1$, we can get:
\begin{equation}\label{equ:28}
\small
\begin{split}
&\mathbb{E} \sum_{k=1}^{N} p_k ||\overline{w}_t - w _t^k ||^2 \\
&= \mathbb{E} \sum_{k=1}^{N} p_k || (w_t^k - \overline{w}_{t_0}) - (\overline{w}_t - \overline{w}_{t_0}) ||^2 \\
&\leq \mathbb{E} \sum_{k=1}^{N} p_k || w_t^k - \overline{w}_{t_0}||^2 \\
&\leq \sum_{k=1}^{N} p_k \mathbb{E} \sum_{t = t_0}^{t - 1} (T - 1) \eta_t^2 || \nabla \mathcal{F}_k (w_t^k, \xi_t^k) + \nabla \mathcal{G}_k (w_t^k, \delta_t^k) ||^2 \\
&\leq \sum_{k=1}^{N} p_k \mathbb{E} \sum_{t = t_0}^{t - 1} (T - 1) \eta_t^2 2 \Big[ || \nabla \mathcal{F}_k (w_t^k, \xi_t^k) ||^2  + || \nabla \mathcal{G}_k (w_t^k, \delta_t^k) ||^2 \Big] \\
&\leq 2 \sum_{k=1}^{N} p_k \sum_{t = t_0}^{t - 1} (T - 1) \eta_{t_0}^2 (G_1^2 + G_2^2) \\
&\leq 2 \sum_{k=1}^{N} p_k \eta_{t_0}^2 (T - 1)^2 (G_1^2 + G_2^2) \\
&\leq 8\eta_t^2 (T - 1)^2 (G_1^2 + G_2^2).
\end{split}
\end{equation}

\begin{table*}[htbp]
\caption{IID and non-IID device data settings for MNIST, CIFAR-10, and CIFAR-100}\label{tab:dist}
\centering
\resizebox{0.75\linewidth}{!}{
\begin{tabular}{c||c|c|c}
\toprule[2pt]
Dataset                  & MNIST & CIFAR-10 & CIFAR-100 \\ \hline
Training Sample \# in Total       & 60000  & 50000    & 50000     \\ \hline
Training Sample \# per AIoT Device & 600   & 500      & 500       \\ \hline
Label \# in Total        & 10     & 10       & 20        \\ \midrule[1pt]
Scenario            & \multicolumn{3}{c}{Data Setting}                                                                                             \\ \hline
IID                  & \multicolumn{3}{c}{Uniform distribution}                                                                                     \\ \hline
non-IID                  & \multicolumn{3}{c}{\begin{tabular}[c]{@{}c@{}}80\% belong to one label,\\ the remaining 20\% belong to other labels\end{tabular}} \\ \bottomrule[2pt]
\end{tabular}
}
\end{table*}

Therefore, we can obtain Formula~\ref{equ:29} by combining Formulas~\ref{equ:a13},~\ref{equ:27} and~\ref{equ:28}.
\begin{equation}\label{equ:29}
\begin{split}
\mathbb{E} || \overline{w}_{t+1} - w^\star ||^2 \leq (1 - 2 \mu \eta_t) \mathbb{E} || \overline{w}_t - w^\star ||^2 + \eta_t^2 B, \\
B = 32 (T - 1)^2 (G_1^2 + G_2^2) \sum_{k = 1}^N p_k^2 (\alpha_k^2 + \beta_k^2) + 8 L \Gamma.
\end{split}
\end{equation}

For a diminishing stepsize similar to~\cite{convergence}, $\eta_t = \frac{\beta}{t + \gamma}$ for some $\beta > \frac{1}{\mu}$ and $\gamma > 0$ such that $\eta_1 \leq min \{\frac{1}{\mu}, \frac{1}{4L}\} = \frac{1}{4L}$ and $\eta_t \leq 2 \eta_{t+T}$.
We will prove $\Delta_t \leq \frac{v}{\gamma + t}$ where $v = max \Big\{\frac{\beta^2 B}{2 \beta \mu - 1}, (\gamma + 1) \Delta_1 \Big\}$.

We prove it by induction. 
Firstly, the definition of $v$ ensures that it holds for $t = 1$. 
Assuming the conclusion holds for some $t$, it follows that:
\begin{equation}
\begin{split}
\mathbb{E} || \overline{w}_{t+1} - w^\star ||^2 &\leq (1 - \frac{2 \beta \mu}{t + \theta}) \frac{v}{t + \theta} + \frac{\beta^2 B}{(t + \theta)^2} \\
&= \frac{t + \theta - 1}{(t + \theta)^2} v + \Big[\frac{\beta^2 B}{(t + \theta)^2} -\frac{2\beta \mu - 1}{(t + \theta)^2} v \Big] \\
&\leq \frac{v}{t + \theta + 1}.
\end{split}
\end{equation}

Then, by the L-smoothness of $\Phi(\cdot)$, we can get:
\begin{equation}\label{equ:31}
\mathbb{E} \Big[\Phi(\overline{w}_t) \Big] - \Phi^\star \leq \frac{L}{2} \mathbb{E} || \overline{w}_t - w^\star ||^2 \leq \frac{L}{2} \frac{v}{\theta + t}.
\end{equation}
where
\begin{equation}\label{equ:32}
v = max \Big\{\frac{\beta^2 B}{2 \beta \mu - 1}, (\gamma + 1) \Delta_1 \Big\}.
\end{equation}
and
\begin{equation}\label{equ:33}
B = 32 (T - 1)^2 (G_1^2 + G_2^2) \sum_{k = 1}^N p_k^2 (\alpha_k^2 + \beta_k^2) + 8 L \Gamma.
\end{equation}

Therefore, our DFL converges to the global optimum at a rate of $O(\frac{1}{t})$ for strongly convex and smooth functions.
For the case of partial device participation, similar to~\cite{convergence}, we can claim that the convergence rate of partial device participation is the same as that of full device participation.

\section{Experimental Results}\label{sec:case}

\subsection{Experimental Setup}

To evaluate the effectiveness of our DFL approach, we implemented the approach on top of a cloud-based architecture consisting of a cloud server and a series of AIoT devices.
Our DFL architecture was built on a workstation (with Intel i7-9700k CPU, 64GB memory, NVIDIA GeForce GTX 2080Ti GPU), and ten Nvidia Jetson Nano boards (with ARM Cortex-A57 processor and 4 GB memory).
Note that in the experiment, only $10$ of the AIoT devices were emulated by the Jetson Nano boards, while the remaining devices were simulated on the workstation. 
The Jetson Nano boards connect to the workstation via a WiFi environment.
Since not all devices are able to participate in each round of model training in the real AIoT application scenario, we set the fraction of AIoT devices to $C=0.1$, i.e., $10$ devices were randomly selected to participate in each round of model training.  
For each AIoT device, we set the batch size, learning rate, and epoch of local training to $50$, $0.01$, and $5$, respectively.
For the performance comparisons of five methods, we set the threshold $\mathcal{T}=0.6$ as an empirical optimal choice, which is detailed in Section~\ref{exp:ratio}.  
Note that for the other hyperparameters of each baseline, we follow the parameters provided by the paper authors.

We conducted experiments on four well-known benchmarks, i.e., MNIST, CIFAR-10, CIFAR-100~\cite{data} and FEMNIST~\cite{leaf}, respectively.
In the experiments, we assumed that there are $100$ AIoT devices for the first three benchmarks, respectively.
Considering that all the AIoT devices are memory limited, we set the training samples of each benchmark equally to all the AIoT devices while putting the $10000$ test samples in the cloud server.
In order to verify the model performance for different data distributions, we set two data scenarios (i.e., the IID scenario and the non-IID scenario) shown in Table~\ref{tab:dist} based on the Dirichlet Distribution according to~\cite{measuring}.
For the IID scenario, all data samples were uniformly distributed on all the $100$ AIoT devices.
For the non-IID scenario, we set that $80\%$ of the data samples on each device belong to one label, while the other $20\%$ belong to other labels evenly.
Note that the CIFAR-100 dataset has two types of sample labels, i.e., the fine-grained label (100 classes) and the coarse-grained label (20 superclasses).  
According to the settings of our experimental scenario, we chose the coarse-grained labels as the sample categories to better distinguish the performance of different methods.
For the dataset FEMNIST from LEAF, we considered a non-IID scenario with $180$ AIoT devices, where each device consists of more than 100 local samples~\footnote{Using the command: ./preprocess.sh -s niid --sf 0.05 -k 100 -t sample}.
Note that the raw data of FEMNIST is naturally non-IID distributed, involving class imbalance, data imbalance, and data heterogeneity.

\begin{figure*}[htbp]
\centering
\subfigure[MNIST]{
\includegraphics[width=0.32\linewidth]{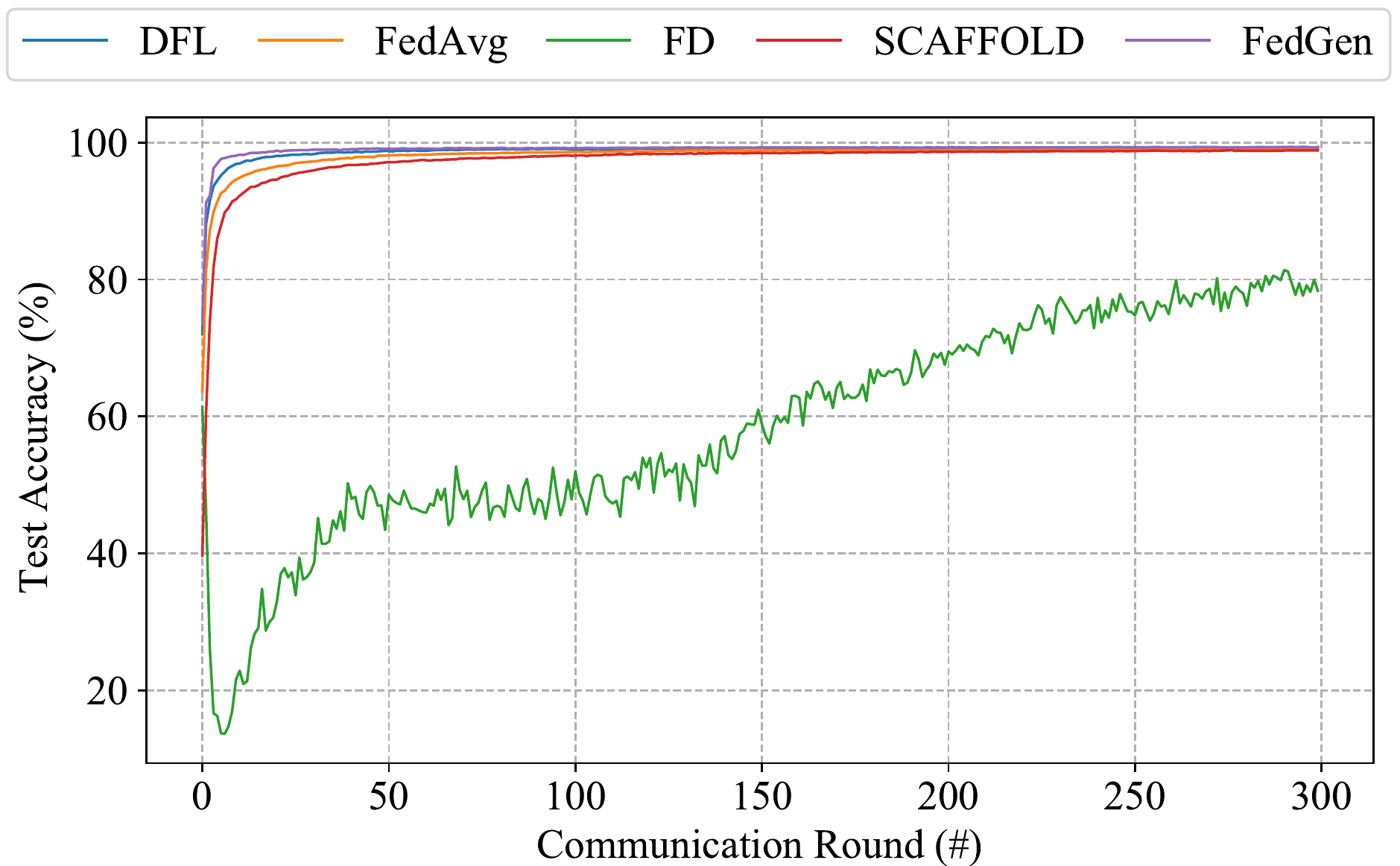}}
\subfigure[CIFAR-10]{
\includegraphics[width=0.32\linewidth]{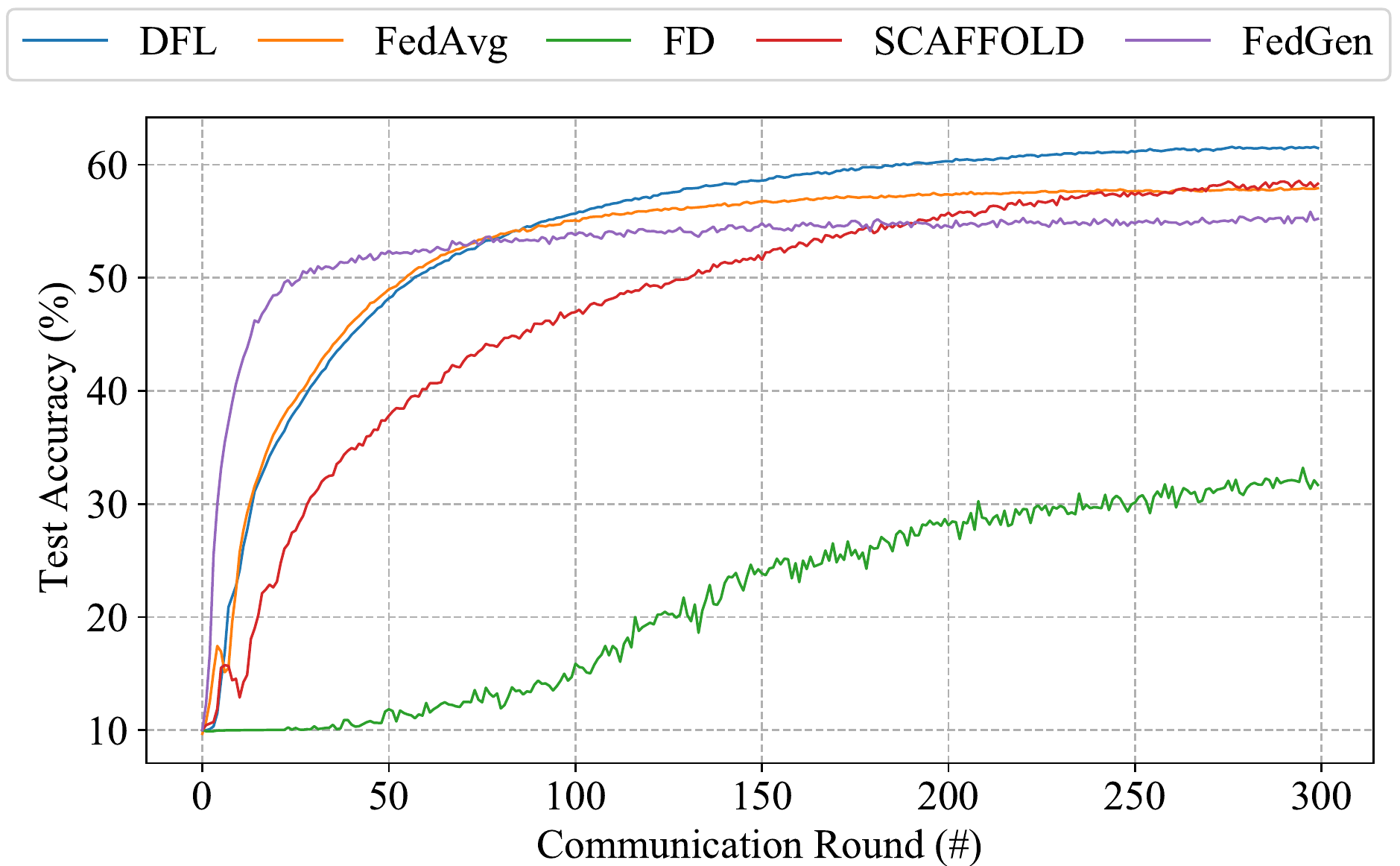}}
\subfigure[CIFAR-100]{
\includegraphics[width=0.32\linewidth]{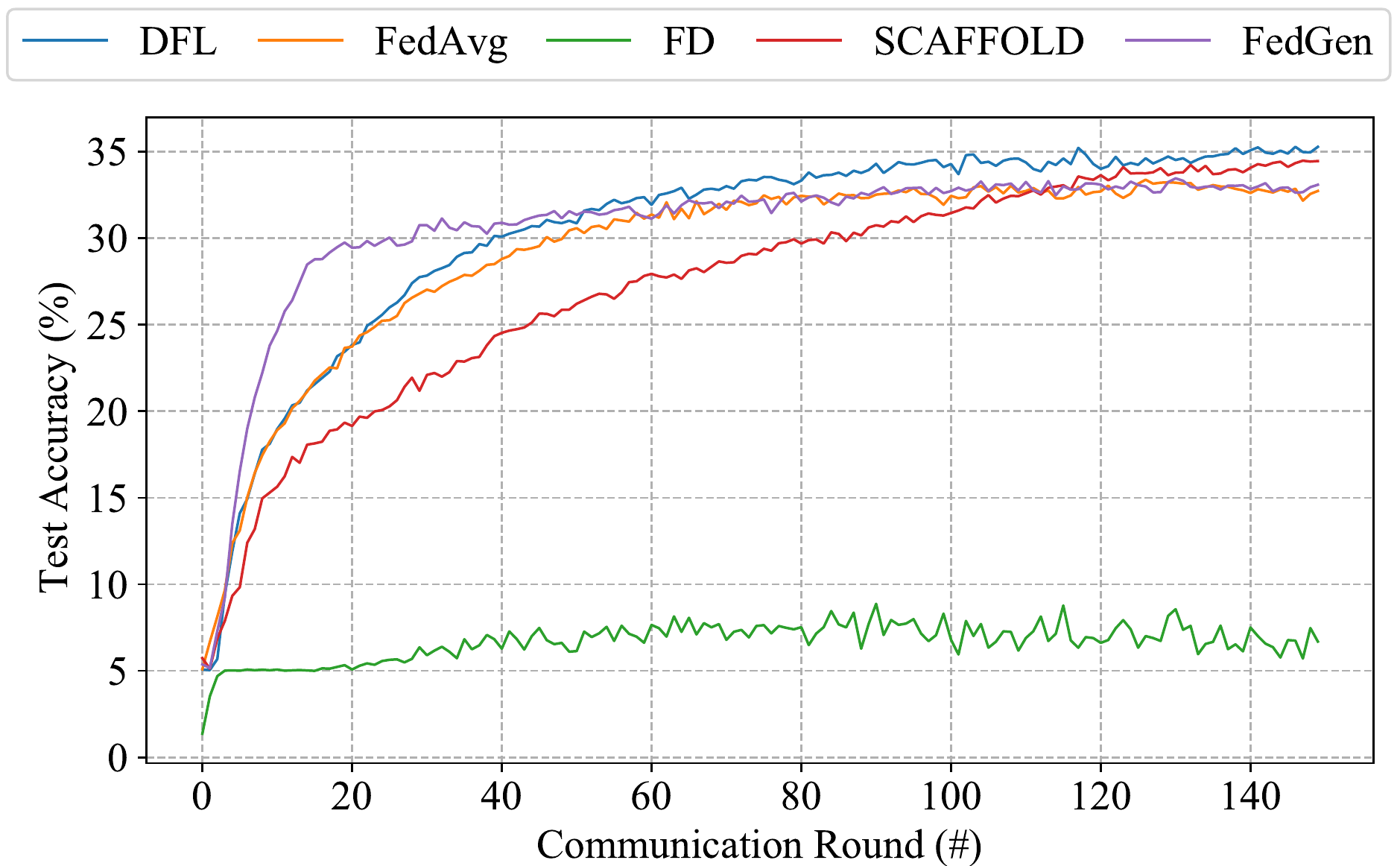}}

\caption{Test accuracy comparison for the IID scenario using CNN}\label{fig:iid}
\end{figure*}

\begin{table*}[htbp]

\caption{Test accuracy comparison for the IID scenario using four models}\label{tab:iid}
\centering
\resizebox{0.8\linewidth}{!}{
\begin{tabular}{|c|c||ccccc|}

\hline
\multirow{2}{*}{Dataset}   & \multirow{2}{*}{Model} & \multicolumn{5}{c|}{Test Accuracy of Different Methods(\%)}                                                                                           \\ \cline{3-7} 
                           &                        & \multicolumn{1}{c|}{FedAvg} & \multicolumn{1}{c|}{FD}    & \multicolumn{1}{c|}{SCAFFOLD}       & \multicolumn{1}{c|}{FedGen}        & DFL (Ours)            \\ \hline \hline
\multirow{4}{*}{MNIST}     & CNN                    & \multicolumn{1}{c|}{99.08}  & \multicolumn{1}{c|}{78.36} & \multicolumn{1}{c|}{98.84}          & \multicolumn{1}{c|}{99.24}          & \textbf{99.33} \\ \cline{2-7} 
                           & ResNet-20              & \multicolumn{1}{c|}{97.86}  & \multicolumn{1}{c|}{74.68} & \multicolumn{1}{c|}{98.22}          & \multicolumn{1}{c|}{98.91} & \textbf{98.93}          \\ \cline{2-7} 
                           & VGG-16                 & \multicolumn{1}{c|}{99.13}  & \multicolumn{1}{c|}{88.40} & \multicolumn{1}{c|}{98.79}          & \multicolumn{1}{c|}{99.19}          & \textbf{99.38} \\ \cline{2-7} 
                           & MobileNetV2            & \multicolumn{1}{c|}{98.96}  & \multicolumn{1}{c|}{11.36} & \multicolumn{1}{c|}{\textbf{99.35}} & \multicolumn{1}{c|}{99.18}          & 99.23          \\ \hline
\multirow{4}{*}{CIFAR-10}  & CNN                    & \multicolumn{1}{c|}{57.92}  & \multicolumn{1}{c|}{31.69} & \multicolumn{1}{c|}{58.32}          & \multicolumn{1}{c|}{55.22}          & \textbf{61.48} \\ \cline{2-7} 
                           & ResNet-20              & \multicolumn{1}{c|}{63.06}  & \multicolumn{1}{c|}{30.11} & \multicolumn{1}{c|}{62.99}          & \multicolumn{1}{c|}{63.35}          & \textbf{64.18} \\ \cline{2-7} 
                           & VGG-16                 & \multicolumn{1}{c|}{79.81}  & \multicolumn{1}{c|}{33.39} & \multicolumn{1}{c|}{81.63}          & \multicolumn{1}{c|}{80.27}          & \textbf{82.30} \\ \cline{2-7} 
                           & MobileNetV2            & \multicolumn{1}{c|}{65.45}  & \multicolumn{1}{c|}{11.52} & \multicolumn{1}{c|}{66.82}          & \multicolumn{1}{c|}{65.67}          & \textbf{69.64} \\ \hline
\multirow{4}{*}{CIFAR-100} & CNN                    & \multicolumn{1}{c|}{32.73}  & \multicolumn{1}{c|}{6.69}  & \multicolumn{1}{c|}{34.46}          & \multicolumn{1}{c|}{33.09}          & \textbf{35.28} \\ \cline{2-7} 
                           & ResNet-20              & \multicolumn{1}{c|}{42.65}  & \multicolumn{1}{c|}{18.62} & \multicolumn{1}{c|}{42.86}          & \multicolumn{1}{c|}{41.66}          & \textbf{43.16} \\ \cline{2-7} 
                           & VGG-16                 & \multicolumn{1}{c|}{55.21}  & \multicolumn{1}{c|}{5.06}  & \multicolumn{1}{c|}{55.39}          & \multicolumn{1}{c|}{55.73}          & \textbf{56.10} \\ \cline{2-7} 
                           & MobileNetV2            & \multicolumn{1}{c|}{41.76}  & \multicolumn{1}{c|}{15.69} & \multicolumn{1}{c|}{41.28}          & \multicolumn{1}{c|}{40.83}          & \textbf{42.85} \\ \hline
                           
\end{tabular}
}
\end{table*}

To fairly validate the effectiveness of different methods, we conducted experiments using four randomly initialized models, i.e., CNN models used in~\cite{communication}, and three popular models (ResNet-20, VGG-16, and MobileNetV2) from Torchvision~\cite{torchvision}. 
The authors in~\cite{communication} designed CNN models for MNIST and CIFAR-10.
For the FEMNIST dataset, we modified the output of the MNIST CNN model to $62$, which is the labels of the samples.
For the CIFAR-100 dataset, we modified the output of the CIFAR-10 CNN model to $20$, which is the coarse-grained labels of the samples.
The Torchvision platform can provide the corresponding model interfaces according to the benchmarks we set.
Therefore, the structure of these three models was fine-tuned according to different benchmarks.

The following sub-sections firstly compare the performance of our proposed DFL with the state-of-the-art methods (i.e., FedAvg~\cite{communication}, FD~\cite{fd},  SCAFFOLD~\cite{scaffold} and FedGen~\cite{datafree}).
Then, we investigate the impact of the loss function ratio and find the empirical optimal ratio of the two-loss functions with a series of experiments.
To avoid the interference of random model initialization and out-of-order dataset training on the experimental results, we ran each experiment ten times and took its mean value for a fair comparison.

\subsection{Performance Evaluation}

\subsubsection{Performance Comparison for IID Scenarios}\label{sec:iid}

In the first experiment, we compared the performance of our method with four baseline methods using the IID scenario set in Table~\ref{tab:dist}.
During the model training process of all the five methods, we tested the inference accuracy of the global models after each round of model aggregation in the cloud server. 
The model accuracy is equal to the ratio of the correctly predicted samples over the total testing samples using the cloud aggregated model.
Due to the space limitation, we show the model accuracy trends using the CNN model on three benchmarks (i.e., MNIST, CIFAR-10, CIFAR-100) along with the number of training rounds in Figure~\ref{fig:iid}.
For each figure, the X-axis denotes the number of training rounds, and the Y-axis indicates the model accuracy.  
Five curves with different colors represent the trends of the model inference accuracy of five different methods.  
From Figure~\ref{fig:iid}, we can find that the model accuracy of all the methods improves with the increase of training rounds.
When the model accuracy does not increase significantly, we believe that the model converges.
Since the FD method converges difficultly, we adaptively present the model convergence process of other methods in Figure~\ref{fig:iid}.

\begin{figure*}[htbp]
\centering
\subfigure[MNIST]{
\includegraphics[width=0.42\linewidth]{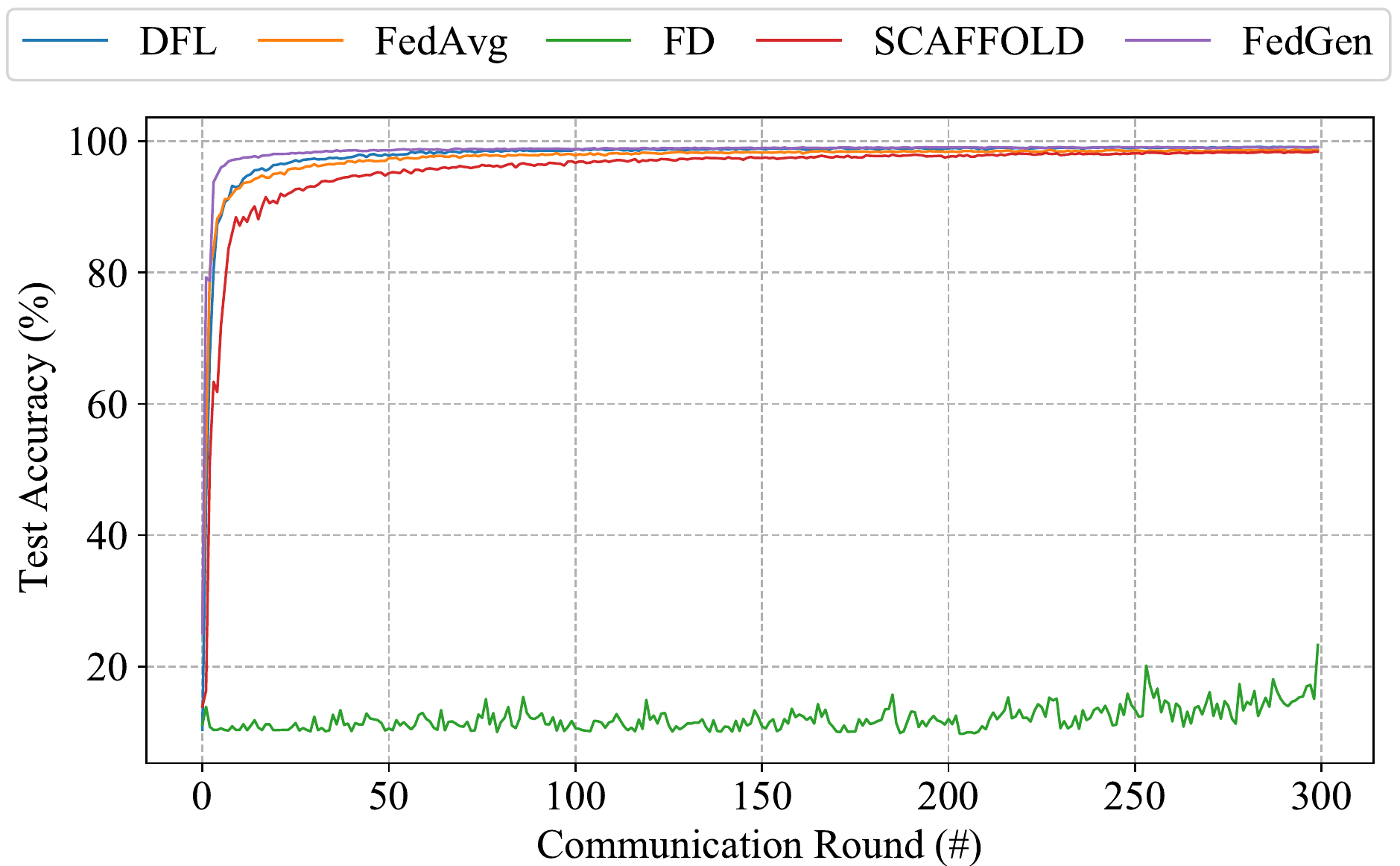}}
\subfigure[CIFAR-10]{
\includegraphics[width=0.42\linewidth]{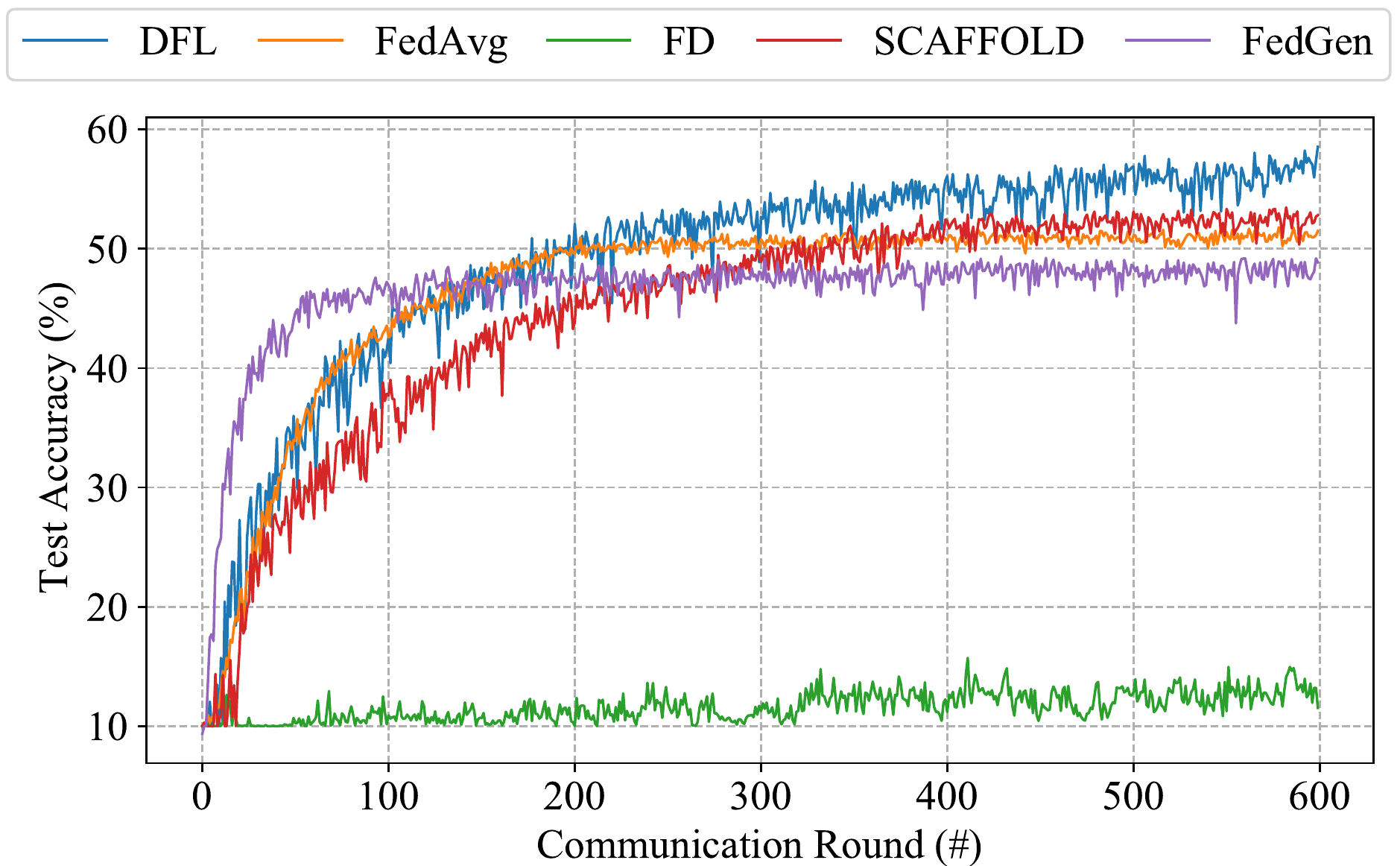}}

\subfigure[CIFAR-100]{
\includegraphics[width=0.42\linewidth]{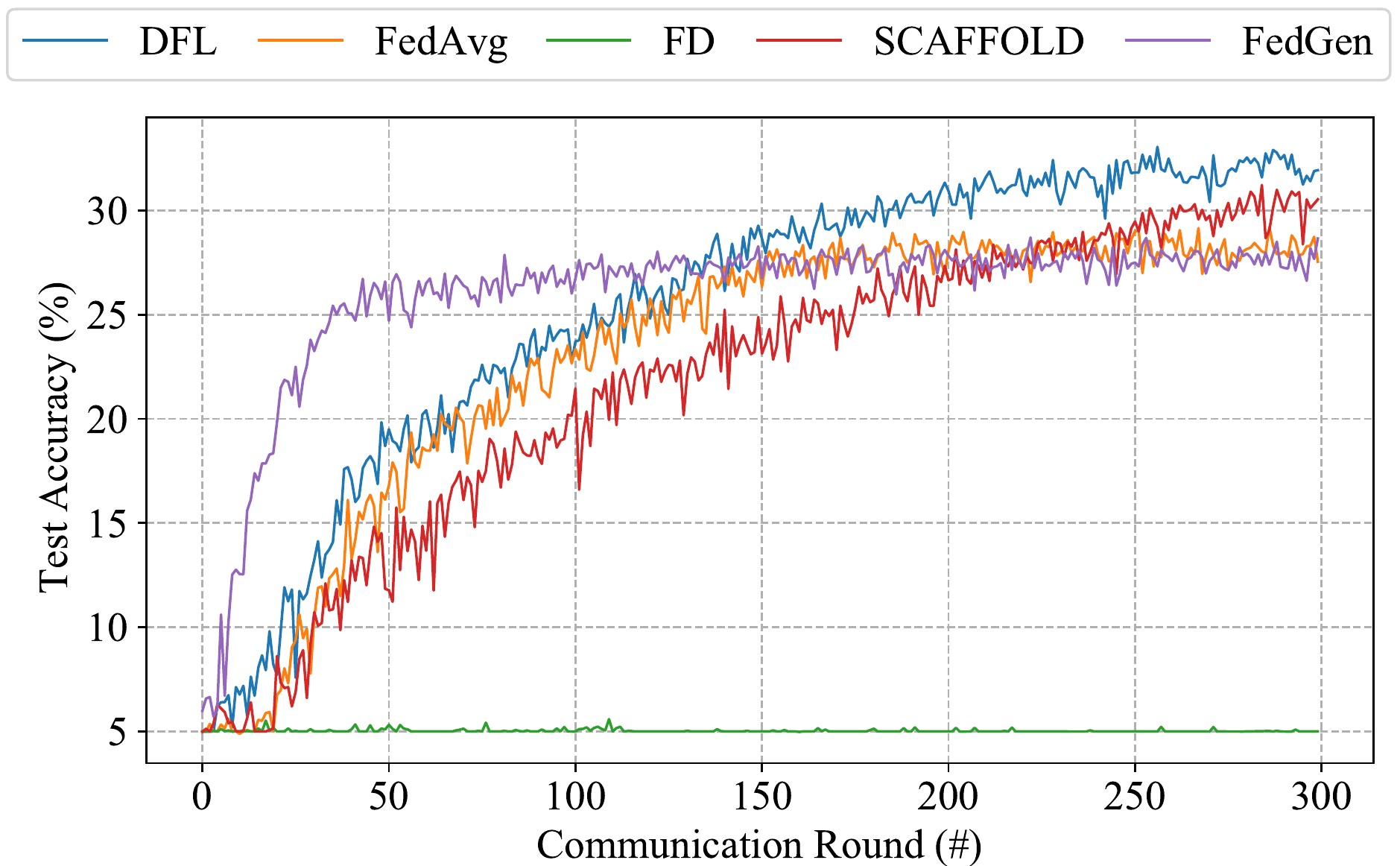}}
\subfigure[FEMNIST]{
\includegraphics[width=0.42\linewidth]{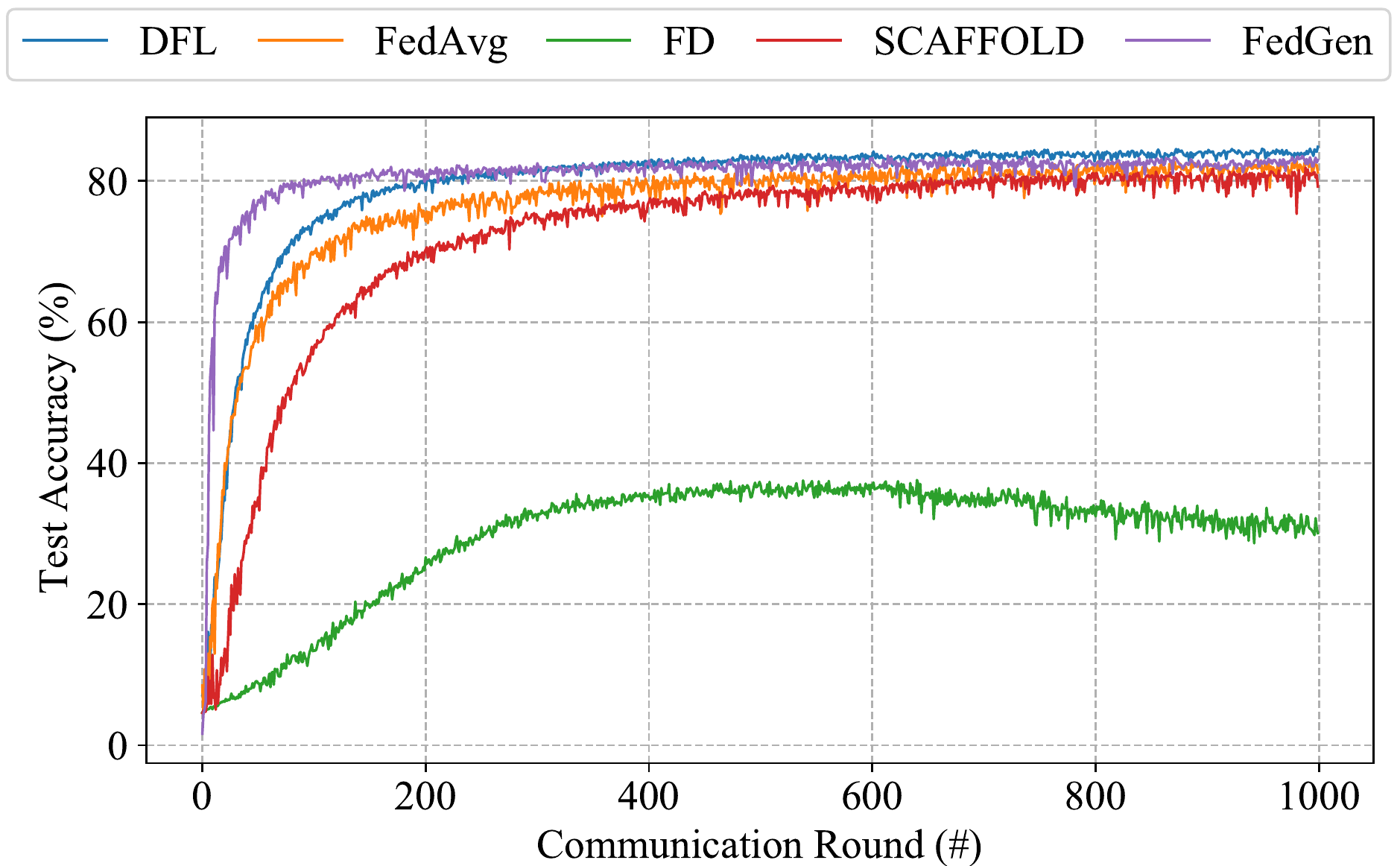}}

\caption{Test accuracy comparison for the non-IID scenario using CNN}\label{fig:noniid}
\end{figure*}

From Figure~\ref{fig:iid}, we can find that our DFL method achieves the highest model accuracy compared with the other four methods on all three benchmarks.
Since our dynamic adjustment strategy gives the soft targets a small proportion in the early stage of model training, we can greatly reduce the side effects on model convergence caused by the insufficient knowledge of soft targets.
We increase the proportion of the soft targets along with the training process, which can improve the model accuracy by maximizing the knowledge of soft targets.
Therefore, our method can effectively improve the model inference accuracy without slowing down the model convergence rate.  
The model accuracy improvement of SCAFFOLD for the IID scenario is insignificant, and the model convergence speed of SCAFFOLD slows down.
This is mainly because the added randomly-initialized global variable misleads the optimization direction of the model in the early stage of model training.
FedGen uses its built-in generators to generate extra samples, thereby speeding up model training.
However, the samples generated by the generators of FedGen is naive, which will decrease the model accuracy in the late stage of the model training.

Table~\ref{tab:iid} presents the complete experimental results of the model accuracy of five methods.
We tested the model accuracy of all the methods using four models on three benchmarks, and the highest model accuracy with the same model for the same dataset is bolded.
From Table~\ref{tab:iid}, we can find our DFL method achieves the highest model inference accuracy in 11 out of 12 cases.
For example, when training the CNN model on the CIFAR-10 dataset, the inference accuracy of FedAvg is $57.92\%$, while SCAFFOLD can achieve $58.32\%$, FedGen can achieve $55.22\%$, and our DFL can achieve $61.48\%$.
This is mainly because the soft targets added by our method can improve the model inference accuracy effectively by enhancing the model knowledge.
The global variables added in the SCAFFOLD method are based on the data distribution relationships among the AIoT devices to guide the model optimization direction of each AIoT device.
Therefore, this method does not greatly improve the model inference accuracy for the IID scenario.
Since the knowledge of the soft targets of FD is less than that of the model gradient, the FD model accuracy is lower than the FedAvg model accuracy.
Note that the generators of FedGen can only generate simple data.
Therefore, the model accuracy of this method becomes worse as the dataset becomes more complex.

\subsubsection{Performance Comparison for Non-IID Scenarios}\label{sec:noniid}

\begin{table*}[htbp]
\caption{Test accuracy comparison for the non-IID scenario using four models}\label{tab:noniid}
\centering
\resizebox{0.8\linewidth}{!}{
\begin{tabular}{|c|c||ccccc|}
\hline
\multirow{2}{*}{Dataset}   & \multirow{2}{*}{Model} & \multicolumn{5}{c|}{Test Accuracy of Different Methods(\%)}                                                                                     \\ \cline{3-7} 
                           &                        & \multicolumn{1}{c|}{FedAvg} & \multicolumn{1}{c|}{FD}    & \multicolumn{1}{c|}{SCAFFOLD} & \multicolumn{1}{c|}{FedGen}         & DFL (Ours)            \\ \hline
\multirow{4}{*}{MNIST}     & CNN                    & \multicolumn{1}{c|}{98.70}  & \multicolumn{1}{c|}{23.27} & \multicolumn{1}{c|}{98.42}    & \multicolumn{1}{c|}{99.08}          & \textbf{99.12} \\ \cline{2-7} 
                           & ResNet-20              & \multicolumn{1}{c|}{95.91}  & \multicolumn{1}{c|}{35.31} & \multicolumn{1}{c|}{96.07}    & \multicolumn{1}{c|}{96.90}          & \textbf{97.28} \\ \cline{2-7} 
                           & VGG-16                 & \multicolumn{1}{c|}{98.71}  & \multicolumn{1}{c|}{13.24} & \multicolumn{1}{c|}{98.31}    & \multicolumn{1}{c|}{\textbf{99.33}} & 99.26          \\ \cline{2-7} 
                           & MobileNetV2            & \multicolumn{1}{c|}{98.34}  & \multicolumn{1}{c|}{11.05} & \multicolumn{1}{c|}{98.45}    & \multicolumn{1}{c|}{98.44}         & \textbf{98.62} \\ \hline
\multirow{4}{*}{CIFAR-10}  & CNN                    & \multicolumn{1}{c|}{51.48}  & \multicolumn{1}{c|}{11.53} & \multicolumn{1}{c|}{52.81}    & \multicolumn{1}{c|}{48.86}          & \textbf{58.54} \\ \cline{2-7} 
                           & ResNet-20              & \multicolumn{1}{c|}{50.86}  & \multicolumn{1}{c|}{20.55} & \multicolumn{1}{c|}{\textbf{54.29}}    & \multicolumn{1}{c|}{50.64}          & 53.03 \\ \cline{2-7} 
                           & VGG-16                 & \multicolumn{1}{c|}{64.55}  & \multicolumn{1}{c|}{17.12} & \multicolumn{1}{c|}{66.74}    & \multicolumn{1}{c|}{62.46}          & \textbf{72.18} \\ \cline{2-7} 
                           & MobileNetV2            & \multicolumn{1}{c|}{38.01}  & \multicolumn{1}{c|}{10.66} & \multicolumn{1}{c|}{39.66}    & \multicolumn{1}{c|}{30.94}          & \textbf{40.06} \\ \hline
\multirow{4}{*}{CIFAR-100} & CNN                    & \multicolumn{1}{c|}{27.54}  & \multicolumn{1}{c|}{5.59}  & \multicolumn{1}{c|}{30.53}    & \multicolumn{1}{c|}{28.65}          & \textbf{31.93} \\ \cline{2-7} 
                           & ResNet-20              & \multicolumn{1}{c|}{17.62}  & \multicolumn{1}{c|}{7.31}  & \multicolumn{1}{c|}{23.48}    & \multicolumn{1}{c|}{23.92}          & \textbf{31.35} \\ \cline{2-7} 
                           & VGG-16                 & \multicolumn{1}{c|}{33.51}  & \multicolumn{1}{c|}{6.13}  & \multicolumn{1}{c|}{33.37}    & \multicolumn{1}{c|}{32.09}          & \textbf{35.21} \\ \cline{2-7} 
                           & MobileNetV2            & \multicolumn{1}{c|}{17.41}  & \multicolumn{1}{c|}{6.13}  & \multicolumn{1}{c|}{19.26}    & \multicolumn{1}{c|}{22.14}          & \textbf{22.29} \\ \hline
\multirow{4}{*}{FEMNIST}   & CNN                    & \multicolumn{1}{c|}{81.22}  & \multicolumn{1}{c|}{30.09} & \multicolumn{1}{c|}{81.29}    & \multicolumn{1}{c|}{82.56}         & \textbf{84.83} \\ \cline{2-7} 
                           & ResNet-20              & \multicolumn{1}{c|}{76.78}  & \multicolumn{1}{c|}{38.26} & \multicolumn{1}{c|}{75.43}    & \multicolumn{1}{c|}{78.72}          & \textbf{81.93} \\ \cline{2-7} 
                           & VGG-16                 & \multicolumn{1}{c|}{83.50}  & \multicolumn{1}{c|}{26.38} & \multicolumn{1}{c|}{83.07}    & \multicolumn{1}{c|}{82.49}          & \textbf{85.21} \\ \cline{2-7} 
                           & MobileNetV2            & \multicolumn{1}{c|}{81.09} & \multicolumn{1}{c|}{28.55} & \multicolumn{1}{c|}{80.99}    & \multicolumn{1}{c|}{82.10} & \textbf{83.17}          \\ \hline
\end{tabular}
}
\end{table*}

\begin{table*}[htbp]
\caption{Size of models and generators}\label{tab:overhead}
\centering
\resizebox{0.75\linewidth}{!}{
\begin{tabular}{|c||cccc||c|}
\hline
\multirow{2}{*}{Model Name} & \multicolumn{4}{c||}{Model Size (KB)}                                                            & \multirow{2}{*}{Generator Size (KB)} \\ \cline{2-5}
                            & \multicolumn{1}{c|}{MNIST}  & \multicolumn{1}{c|}{CIFAR-10} & \multicolumn{1}{c|}{CIFAR-100} & FEMNIST &                                             \\ \hline
CNN                         & \multicolumn{1}{c|}{265.9}  & \multicolumn{1}{c|}{249.7}    & \multicolumn{1}{c|}{253.1}     & 643.5   & 338.2                                       \\ \hline
ResNet-20                   & \multicolumn{1}{c|}{908.7}  & \multicolumn{1}{c|}{909.9}    & \multicolumn{1}{c|}{1024}      & 962.2   & 691.1                                       \\ \hline
VGG-16                      & \multicolumn{1}{c|}{32768}  & \multicolumn{1}{c|}{137830.4} & \multicolumn{1}{c|}{139366.4}  & 32870.4 & 8806.4                                      \\ \hline
MobileNetV2                 & \multicolumn{1}{c|}{9420.8} & \multicolumn{1}{c|}{9420.8}   & \multicolumn{1}{c|}{9932.8}    & 9728    & 2867.2                                      \\ \hline
\end{tabular}
}
\end{table*}

To evaluate the performance of our DFL method for the non-IID scenario, we compared the five methods (i.e., FedAvg, FD, SCAFFOLD, and FedGen) using four benchmarks (i.e., MNIST, CIFAR-10, CIFAR-100, and FEMNIST), where the former three benchmarks follow the non-IID setting presented in Table~\ref{tab:dist} and the dataset FEMNIST follows the non-IID setting provided by LEAF.
Figure~\ref{fig:noniid} shows the trends of model accuracy using the CNN model along with the number of training rounds.
Similar to the observations from Figure~\ref{fig:iid}, we can find that our approach outperforms the other four methods.
Our DFL method achieves the highest model accuracy and the fastest model convergence speed on all four datasets.

Table~\ref{tab:noniid} presents the complete experimental results of the model accuracy for the non-IID scenario.
From Table~\ref{tab:noniid}, we can find that our DFL method can achieve the best performance in 14 out of 16 cases.
For example, when training the CNN model on the CIFAR-10 dataset, our DFL method outperforms FedAvg, SCAFFOLD and FedGen by $7.06\%$, $5.73\%$ and $9.68\%$, respectively.
The reason why our approach is superior is mainly because the added soft targets can enhance model knowledge, which is effective for both IID and non-IID scenarios.
Therefore, the local training process can use the knowledge of soft targets to improve the model accuracy.
Note that the model accuracy of the FD method is $11.53\%$ for this case, which is similar to that of a randomly initialized model.
This is mainly because the knowledge of soft targets is insufficient to train a model.
Therefore, the model trained by FD using soft targets alone is inaccurate.

\subsubsection{Comparison of Communication Overhead}\label{sec:overhead}

Table~\ref{tab:overhead} illustrates the size of different models as well as the generators used by FedGen, expressed in KB.
The network resources occupied by SCAFFOLD are always twice that of FedAvg due to the additional global variable of each AIoT device.
FedGen needs to dispatch both built-in generators and model gradients, where the size of generators is shown in Table~\ref{tab:overhead}.
Although the information interaction of FD occupies few network resources, the model trained by FD is inaccurate, which makes it unable to be deployed in the AIoT applications.
The communication cost of our DFL method equals the sum of FedAvg and FD since our method adds soft targets based on FedAvg.
The size of soft targets is only determined by the number of categories of datasets (i.e., $3.2$ KB for CIFAR-100 and $0.8$ KB for the other benchmarks).
Therefore, the larger the training model, the smaller the proportion of communication cost increased by our method than FedAvg.
For example, when training the model ResNet-20 on dataset FEMNIST using our DFL approach, the total size of both model gradients and soft targets is $963$ KB, which needs $0.36$ seconds on average for one DFL training round.
However, SCAFFOLD needs $0.71$ seconds for one training round, where the total size of both the global model and the global control variable is $1924.4$ KB.
The total size of both the global model and built-in generators involved in FedGen is $1653.3$ KB, which requires $0.48$ seconds for the interaction between the cloud server and devices.
Compared with the state-of-the-art methods (i.e., SCAFFOLD and FedGen), our DFL method has less communication overhead while trained models can achieve higher accuracy.

\begin{table*}[htbp]
\caption{Test accuracy comparison with different thresholds}\label{tab:threshold}
\resizebox{\linewidth}{!}{
\begin{tabular}{|c|c||c|c|c|c|c|c|c|c|c|c|c|}
\hline
\multirow{2}{*}{Scenario} & \multirow{2}{*}{Model} & \multicolumn{11}{c|}{Test Accuracy (\%)}                                                                         \\ \cline{3-13} 
                          &                        & T=0     & T=0.1   & T=0.2   & T=0.3   & T=0.4   & T=0.5            & T=0.6            & T=0.7   & T=0.8   & T=0.9   & T=1     \\ \hline \hline
\multirow{4}{*}{IID}      & CNN                    & 33.94 & 59.74 & 59.31 & 59.76 & 59.16 & 59.58          & \textbf{61.48} & 60.03 & 60.24 & 59.42 & 57.92 \\ \cline{2-13} 
                          & ResNet-20              & 52.68 & 62.85 & 62.67 & 62.8  & 62.07 & 62.28          & \textbf{64.18} & 63.64 & 63.08 & 61.89 & 63.06 \\ \cline{2-13} 
                          & VGG-16                 & 79.13 & 80.98 & 81.55 & 81.67 & 81.63 & 81.18          & \textbf{82.30}  & 82.23 & 81.24 & 80.12 & 79.81 \\ \cline{2-13} 
                          & MobileNetV2            & 64.09 & 68.96 & 69.01 & 69.41 & 69.59 & \textbf{69.72} & 69.64          & 68.93 & 68.82 & 68.12 & 65.45 \\ \hline
\multirow{4}{*}{non-IID}  & CNN                    & 18.45 & 51.35 & 51.57 & 51.27 & 52.37 & 53.34          & \textbf{58.54} & 54.26 & 55.29 & 54.38 & 51.48 \\ \cline{2-13} 
                          & ResNet-20              & 24.65 & 27.74 & 36.03 & 39.76 & 45.67 & 46.55          & \textbf{53.03} & 52.44 & 50.34 & 51.59 & 50.86 \\ \cline{2-13} 
                          & VGG-16                 & 67.03 & 68.73 & 68.81 & 68.18 & 68.67 & 69.86          & \textbf{72.18} & 68.26 & 68.83 & 66.47 & 64.55 \\ \cline{2-13} 
                          & MobileNetV2            & 33.89 & 35.18 & 34.01 & 37.48 & 38.85 & 39.96          & \textbf{40.06} & 39.82 & 35.88 & 38.34 & 38.01 \\ \hline
\end{tabular}
}
\end{table*}

\subsection{Impacts of Dynamic Adjustment Strategy}\label{exp:ratio}

\begin{figure}[htbp]
	\centering
	\includegraphics[width=0.9\linewidth]{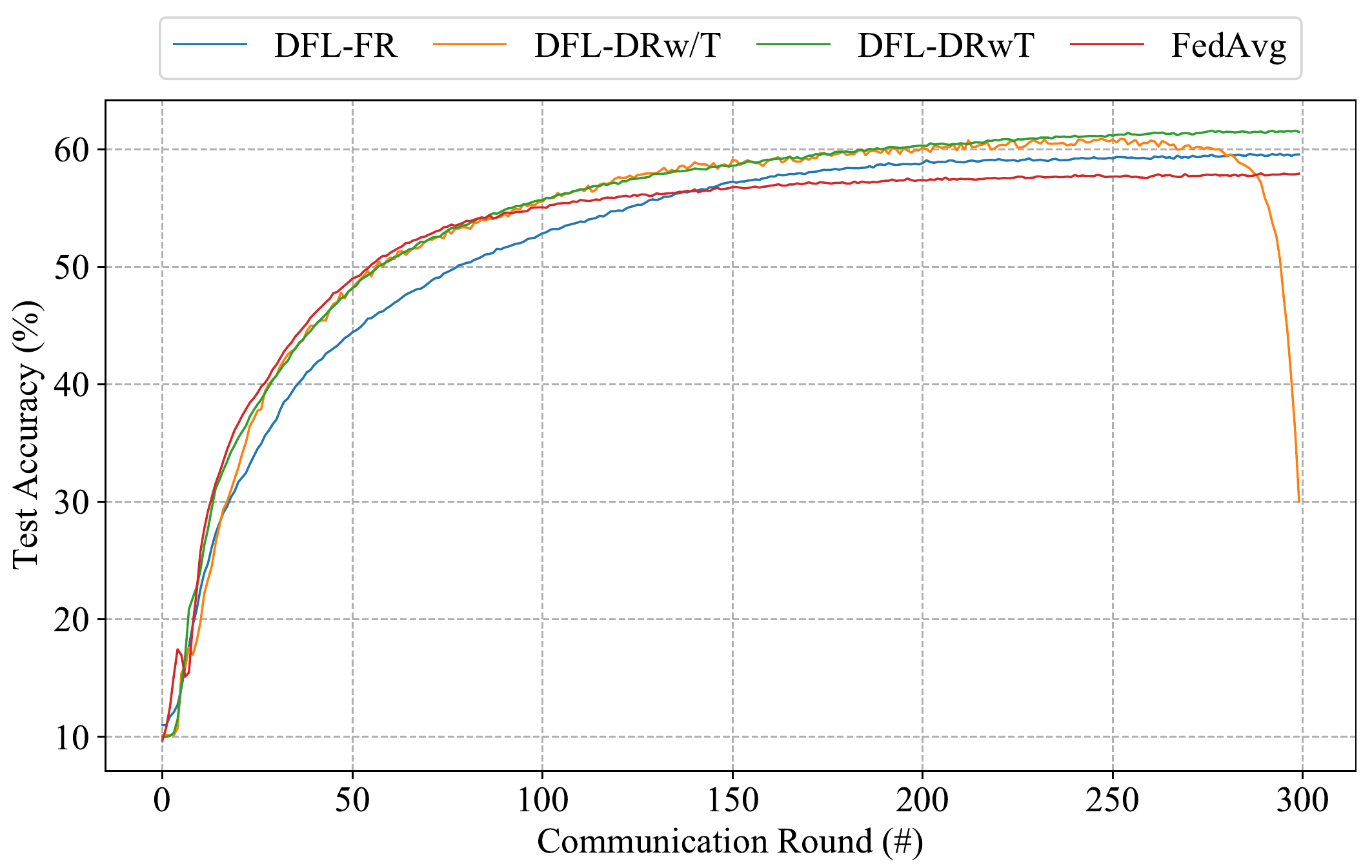}
	\caption{Model accuracy trends of four methods}\label{fig:dynamic}
\end{figure}

Since the ratio (i.e., $\rho$) of the two-loss functions controls the proportion of hard labels and soft targets during the local training, it plays an important role in our DFL approach.
To investigate the impacts of dynamic adjustment strategy, we conducted a series of experiments to verify the role of loss function ratio in different stages of model training.
As a representative, Figure~\ref{fig:dynamic} shows the trends of model accuracy of FedAvg and our DFL method with three different loss function ratio settings using the CNN model for the IID scenario of CIFAR-10.
In Figure~\ref{fig:dynamic}, four curves with different colors represent the model accuracy trends of four methods, i.e., DFL method with Fixed loss function Ratio named DFL-FR (marked in blue), DFL method with Dynamic changing Ratio without the Threshold named DFL-DRw/T (marked in yellow), DFL method with Dynamic changing Ratio and the Threshold named DFL-DRwT (marked in green), and FedAvg (marked in red).

From Figure~\ref{fig:dynamic}, we can find that the model accuracy of DFL-DRwT and DFL-DRw/T increases rapidly in the early stage of training while the DFL-FR model accuracy increases slowly.
This is mainly because the knowledge of soft targets is insufficient in the early stage of model training, which can mislead the model optimization direction.
The knowledge of soft targets increases as the model trains.
In the late stage of training, there is sufficient knowledge of soft targets to guide the model training.
Therefore, assigning a higher proportion to soft targets is beneficial to the model training as the number of training rounds increases.
However, the model accuracy drops sharply in the final stage of the DFL-DRw/T model training. 
This is mainly because the proportion of hard labels is too small, and the model trained with soft targets alone is inaccurate.
Therefore, we need to control the ratio between the two-loss functions to achieve the best model performance after the soft targets gain sufficient knowledge.

To investigate the empirical optimal threshold of the loss function ratio, we conducted experiments with thresholds from $0$ to $1$ with a step length of $0.1$.
Table~\ref{tab:threshold} shows the experimental results of the model accuracy obtained on CIFAR-10 using different thresholds and our two data distribution settings, and the highest model accuracy is bolded.
We can find that our DFL method achieves the highest model accuracy in 7 out of 8 cases when the threshold is set to $0.6$.
Only when MobileNetV2 is used for the IID scenario the model does not achieve the highest accuracy at $\mathcal{T}=0.6$.
In this case, the model obtains the highest accuracy at $\mathcal{T}=0.5$, which is only $0.08$ more than $\mathcal{T}=0.6$.
Therefore, to achieve the best model performance, we set $\mathcal{T}=0.6$ to maximize the use of soft targets and hard labels.

\section{Conclusion and Future work}\label{sec:conclusions}

Although Federated Learning (FL) techniques are becoming popular in Artificial Intelligence Internet of Things (AIoT) applications, they are suffering from the problem of model inaccuracy. 
How to improve the model accuracy of FL under the limited network bandwidth and memory resources is becoming a major bottleneck in the design of AIoT applications.
To address the above problem, this paper presents a novel FL architecture based on Knowledge Distillation (KD) named DFL, which can increase the model generalization ability.
By adding soft targets to each round of model training, our proposed approach can increase the inference accuracy of the FL model without introducing significant communication and memory overhead.
To further improve the performance of our DFL model, we designed a strategy to dynamically adjust the ratio of the two loss functions in KD to maximize the use of knowledge of soft targets.
Comprehensive experimental results on four well-known benchmarks prove the effectiveness of our approach.
For future work, we need to consider a better dynamic adjustment strategy, where the loss function ratio is controlled by the feedback of the model accuracy.



\begin{IEEEbiography}[{\includegraphics[width=1in,height=1.25in,clip,keepaspectratio]{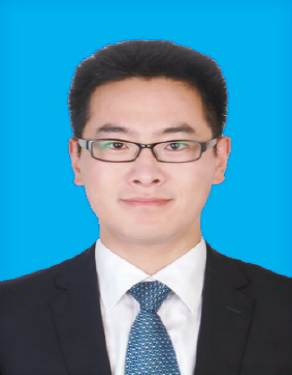}}]{Tian Liu} received the B.S. and M.E. degrees from Department of Computer Science and Technology, Hohai University, Nanjing, China, in 2011 and 2014 respectively, and the Engineer degree from Department of Information and Statistic, Polytech’Lille, France, in 2012. He is currently a Ph.D. student in the Software Engineering Institute, East China Normal University. 
He is also a lecturer in the Department of Information Science and Engineering, Zaozhuang University. His research interests are in the area of federated learning, machine learning, internet of things and cloud computing.
\end{IEEEbiography}

\begin{IEEEbiography}[{\includegraphics[width=1in,height=1.25in,clip,keepaspectratio]{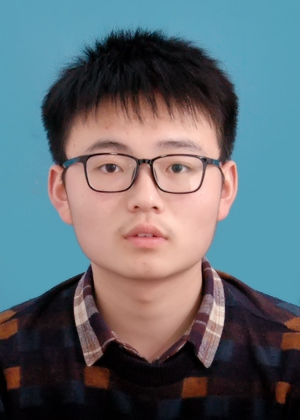}}]{Zhiwei Ling} received the B.S. degrees from Department of Education Information Technology, East China Normal University, Shanghai, China, in 2021. He is currently a Master student in the Software Engineering Institute, East China Normal University. 
His research interests are in the area of federated learning, machine learning,  and internet of things.
\end{IEEEbiography}

\begin{IEEEbiography}[{\includegraphics[width=1in,height=1.25in,clip,keepaspectratio]{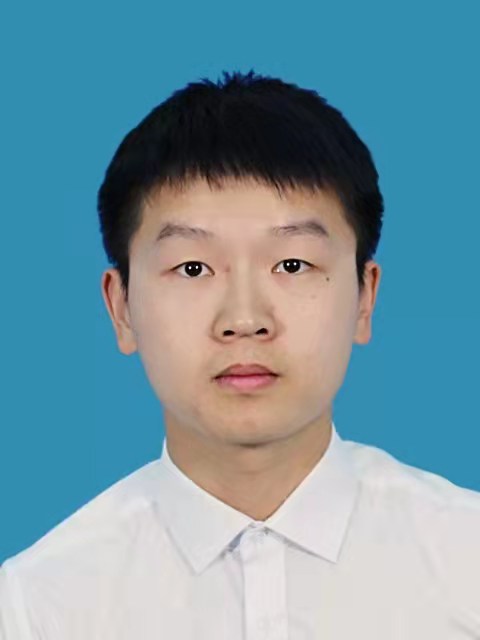}}]{Jun Xia} received the B.S. degree from the Department of Computer Science and Technology, Hainan University, Hainan, China, in 2016 and the M.E. degree from Department of Computer Science and
Technology, Jiangnan University, Wuxi, China in 2019, respectively. He is currently a Ph.D. student in the Software Engineering Institute, East China Normal University, Shanghai, China. His research interests are in the area of federated learning, AIoT applications, cloud computing, and heterogeneous computing.
\end{IEEEbiography}

\begin{IEEEbiography}[{\includegraphics[width=1in,height=1.25in,clip,keepaspectratio]{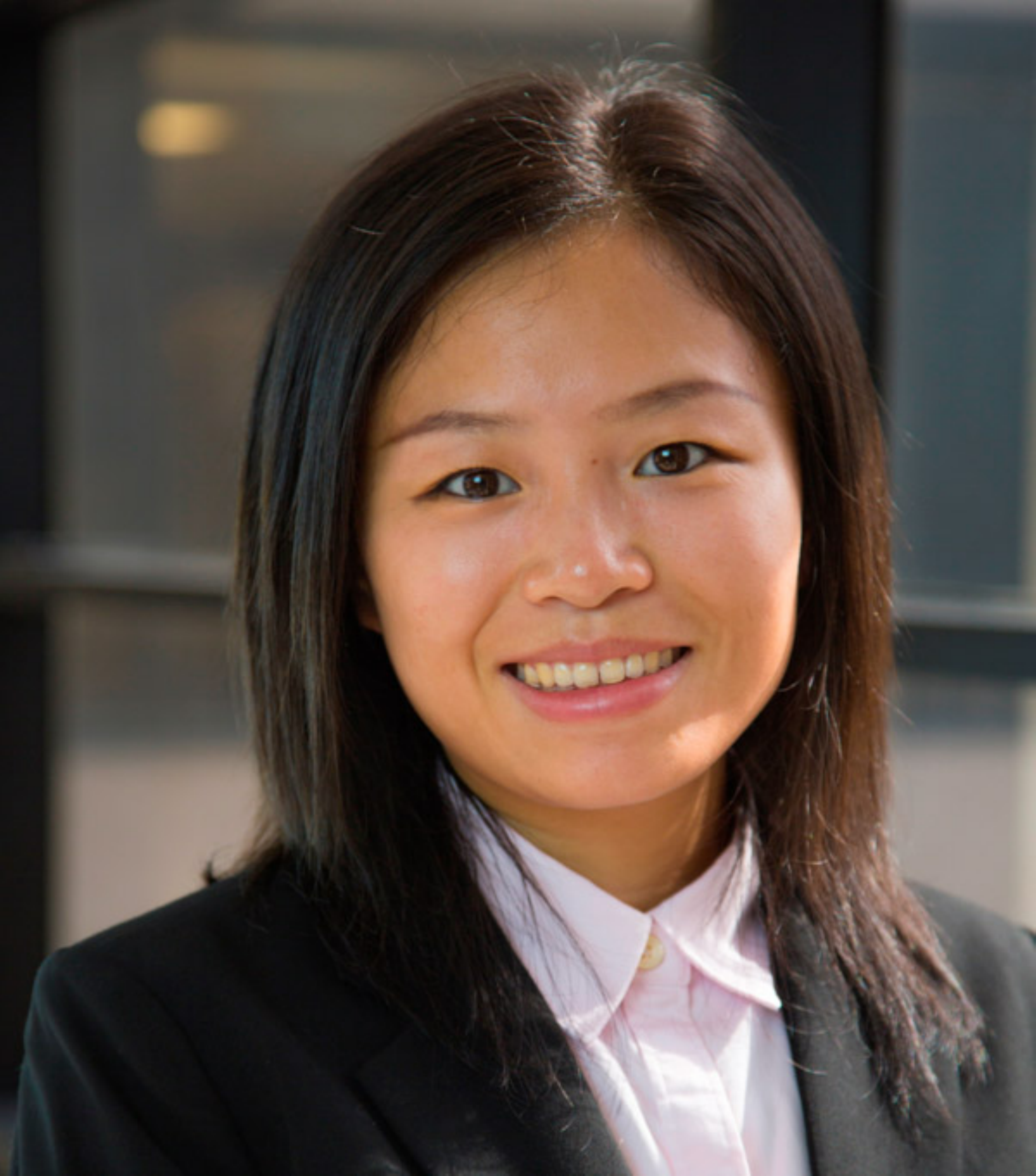}}]{Xin Fu} (SM'10) received the Ph.D. degree in Computer Engineering from the University of Florida, Gainesville, in 2009. She was an NSF Computing Innovation Fellow with the Computer Science Department, the University of Illinois at Urbana-Champaign, Urbana, from 2009 to 2010. From 2010 to 2014, she was an Assistant Professor at the Department of Electrical Engineering and Computer Science, the University of Kansas, Lawrence. Currently, she is an Associate Professor at the Electrical and Computer Engineering Department, the University of Houston, Houston. Her research interests include high-performance computing, machine learning, energy-efficient computing, mobile computing. Dr. Fu is a recipient of 2014 NSF Faculty Early CAREER Award, 2012 Kansas NSF EPSCoR First Award, and 2009 NSF Computing Innovation Fellow.
\end{IEEEbiography}

\begin{IEEEbiography}[{\includegraphics[width=1in,height=1.25in,clip,keepaspectratio]{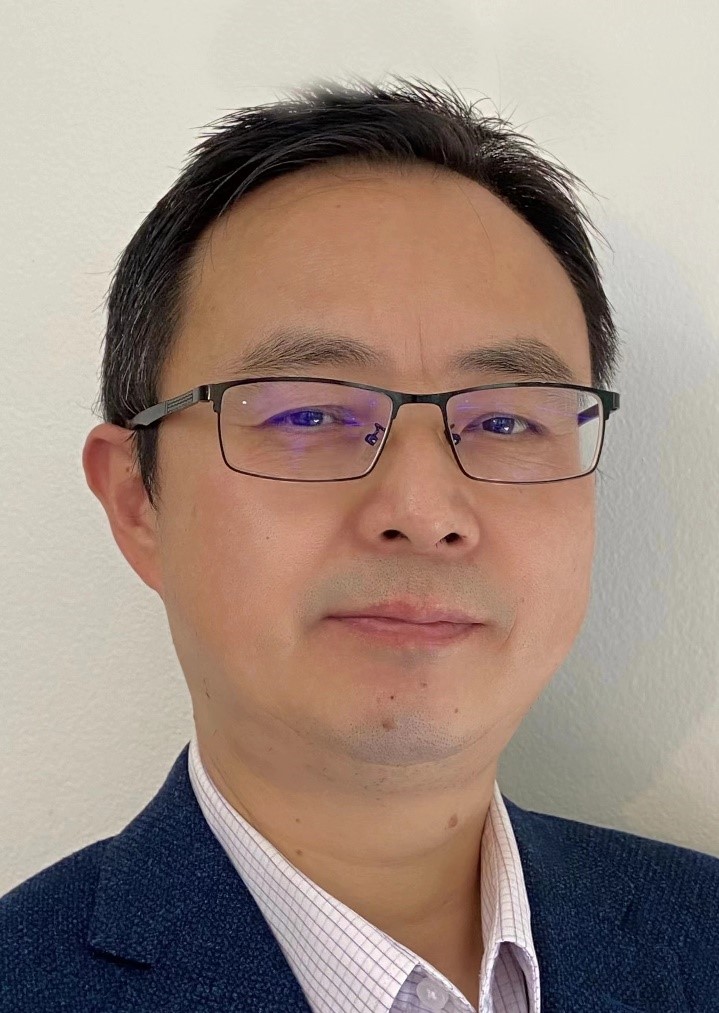}}]{Shui Yu}
(SM’12) obtained his PhD from Deakin University, Australia, in 2004. He currently is a Professor of School of Computer Science, University of Technology Sydney, Australia. Dr Yu’s research interest includes Big Data, Security and Privacy, Networking, and Mathematical Modelling. He has published four monographs and edited two books, more than 400 technical papers, including top journals and top conferences, such as IEEE TPDS, TC, TIFS, TMC, TKDE, TETC, ToN, and INFOCOM. His h-index is 63. Dr Yu initiated the research field of networking for big data in 2013, and his research outputs have been widely adopted by industrial systems, such as Amazon cloud security. He is currently serving a number of prestigious editorial boards, including IEEE Communications Surveys and Tutorials (Area Editor), IEEE Communications Magazine, IEEE Internet of Things Journal, and so on. He served as a Distinguished Lecturer of IEEE Communications Society (2018-2021). He is a Distinguished Visitor of IEEE Computer Society, a voting member of IEEE ComSoc Educational Services board, and an elected member of Board of Governor of IEEE Vehicular Technology Society.
\end{IEEEbiography}

\begin{IEEEbiography}[{\includegraphics[width=1in,height=1.25in,clip,keepaspectratio]{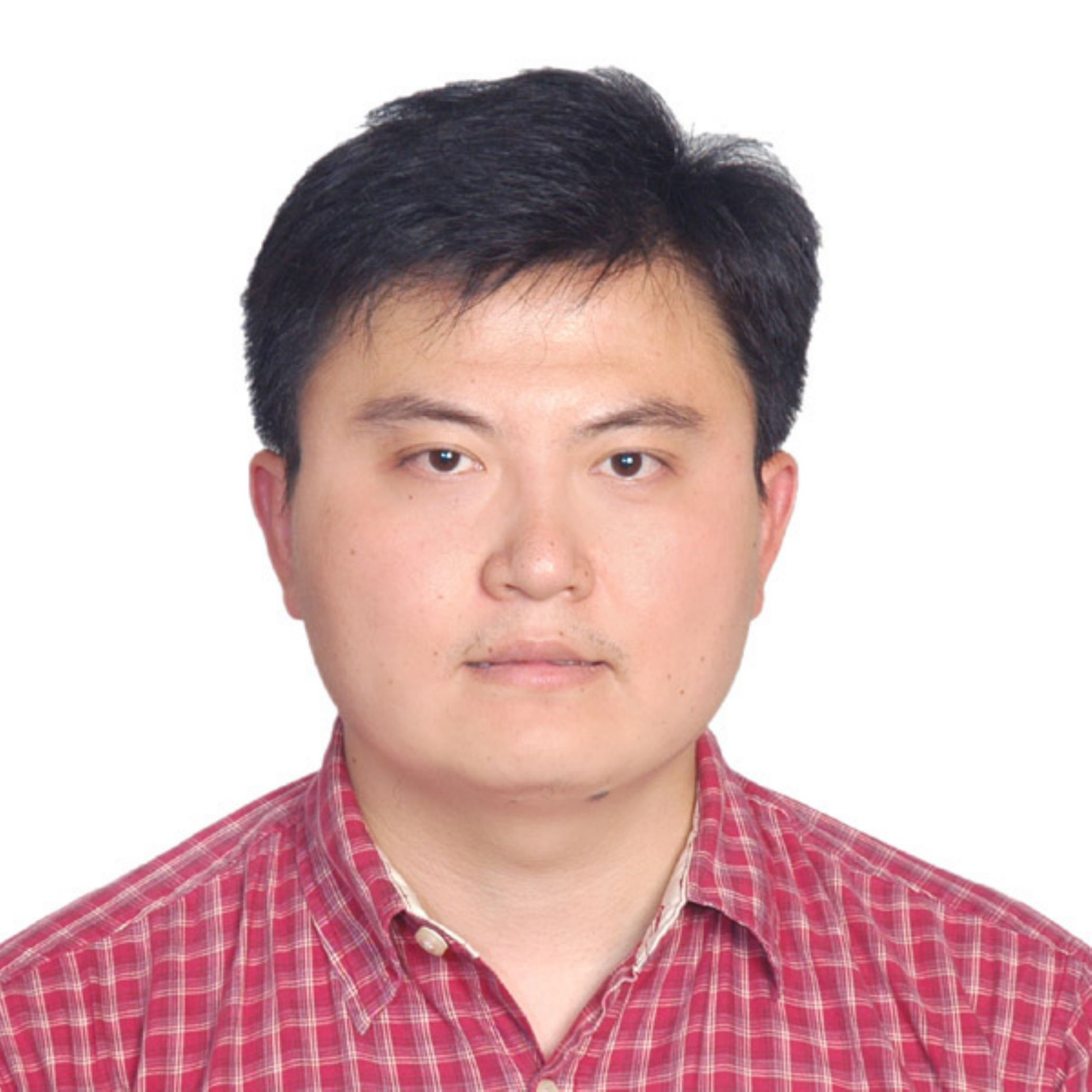}}]{Mingsong Chen}
(M'08--SM'11) received the B.S. and M.E. degrees from Department of Computer Science and Technology, Nanjing University, Nanjing, China, in 2003 and 2006 respectively, and the Ph.D. degree in Computer Engineering from the University of Florida, Gainesville, in 2010. He is currently a Professor with the 
Software Engineering Institute at East China Normal University. His research interests are in the area of cloud computing, design automation of cyber-physical systems, parallel and distributed systems, 
and formal verification techniques. He is an Associate Editor of IET Computers \& Digital Techniques, and Journal of Circuits, Systems and Computers.
\end{IEEEbiography}

\end{document}